\documentclass[10pt,twocolumn,letterpaper]{article}

\usepackage{iccv}
\usepackage{times}
\usepackage{epsfig}
\usepackage{graphicx}
\usepackage{amsmath}
\usepackage{amssymb}
\usepackage{multirow}
\usepackage{booktabs}
\usepackage{pifont}
\usepackage{comment}
\usepackage{amsthm}
\usepackage[utf8]{inputenc}
\usepackage[english]{babel}
\usepackage{makecell}
\usepackage{boldline}
\usepackage{subcaption}
\usepackage[font=small,justification=justified]{caption}
\usepackage[dvipsnames]{xcolor}

\usepackage[pagebackref=true,breaklinks=true,colorlinks,bookmarks=false]{hyperref}

\iccvfinalcopy 

\newtheorem{theorem}{Theorem}
\newtheorem{lemma}{Lemma}

\def\iccvPaperID{1168} 

\ificcvfinal\fi
\begin{document}

\title{MeteorNet: Deep Learning on Dynamic 3D Point Cloud Sequences}

\author{Xingyu Liu\\
Stanford University\\
\and
Mengyuan Yan\\
Stanford University\\
\and
Jeannette Bohg\\
Stanford University\\
}

\definecolor{viz_red}{RGB}{204, 0, 0}
\definecolor{viz_yellow}{RGB}{241, 194, 50}
\definecolor{viz_green}{RGB}{52, 183, 47}

\maketitle

\begin{abstract}
   Understanding dynamic 3D environment is crucial for robotic agents and many other applications. We propose a novel neural network architecture called MeteorNet for learning representations for dynamic 3D point cloud sequences. Different from previous work that adopts a grid-based representation and applies 3D or 4D convolutions, our network directly processes point clouds. We propose two ways to construct spatiotemporal neighborhoods for each point in the point cloud sequence. Information from these neighborhoods is aggregated to learn features per point. We benchmark our network on a variety of 3D recognition tasks including action recognition, semantic segmentation and scene flow estimation. MeteorNet shows stronger performance than previous grid-based methods while achieving state-of-the-art performance on Synthia.  MeteorNet also outperforms previous baseline methods that are able to process at most two consecutive point clouds. To the best of our knowledge, this is the first work on deep learning for dynamic raw point cloud sequences.
\end{abstract}
\vspace{-1ex}
\section{Introduction}

Our world is three dimensional. In many applications such as autonomous driving and robotic manipulation, the autonomous agent needs to understand its 3D environment.
Among various 3D geometric representations, point clouds are the closest to raw sensory data from LiDAR or RGB-D cameras. Point clouds do not suffer as much from quantization errors compared to other geometric representations such as grids.
Recently, deep architectures have been proposed that directly consume a single or a pair of point clouds for various 3D recognition tasks \cite{PointNet, PointNet++,Frustum:PointNet,FlowNet3D,PointRCNN}. These architectures have outperformed methods based on other geometric representations.
 
Our world is also dynamic. Many recognition tasks benefit from sequences of temporal data that are longer than two frames. Examples include estimating the acceleration of a moving object or recognizing human actions. 
Unlike 2D image videos which can be represented as a regular spatiotemporal 3D grid and learned by 3D convolutional neural networks (CNNs) \cite{C3D,Karpathy:Video:CNN,I3D}, dynamic 3D point cloud sequences have an entirely different structure and are more challenging to learn. Recently, deep architectures were proposed for dynamic point cloud sequences that convert the irregular 3D data to a grid representation and leverage 3D or 4D convolutions along the time dimension \cite{FaF,MinkNet}.
However, the grid quantization error is inevitable in these methods and can be fatal for robotic applications that require precise localization. Moreover, performing convolution with a shared and fixed kernel everywhere in the 3D or 4D space is either inefficient or requires special engineering efforts such as sparse convolution \cite{MinkNet}. 

\begin{figure}[t]
\centering
\small
\includegraphics[width=0.9\linewidth]{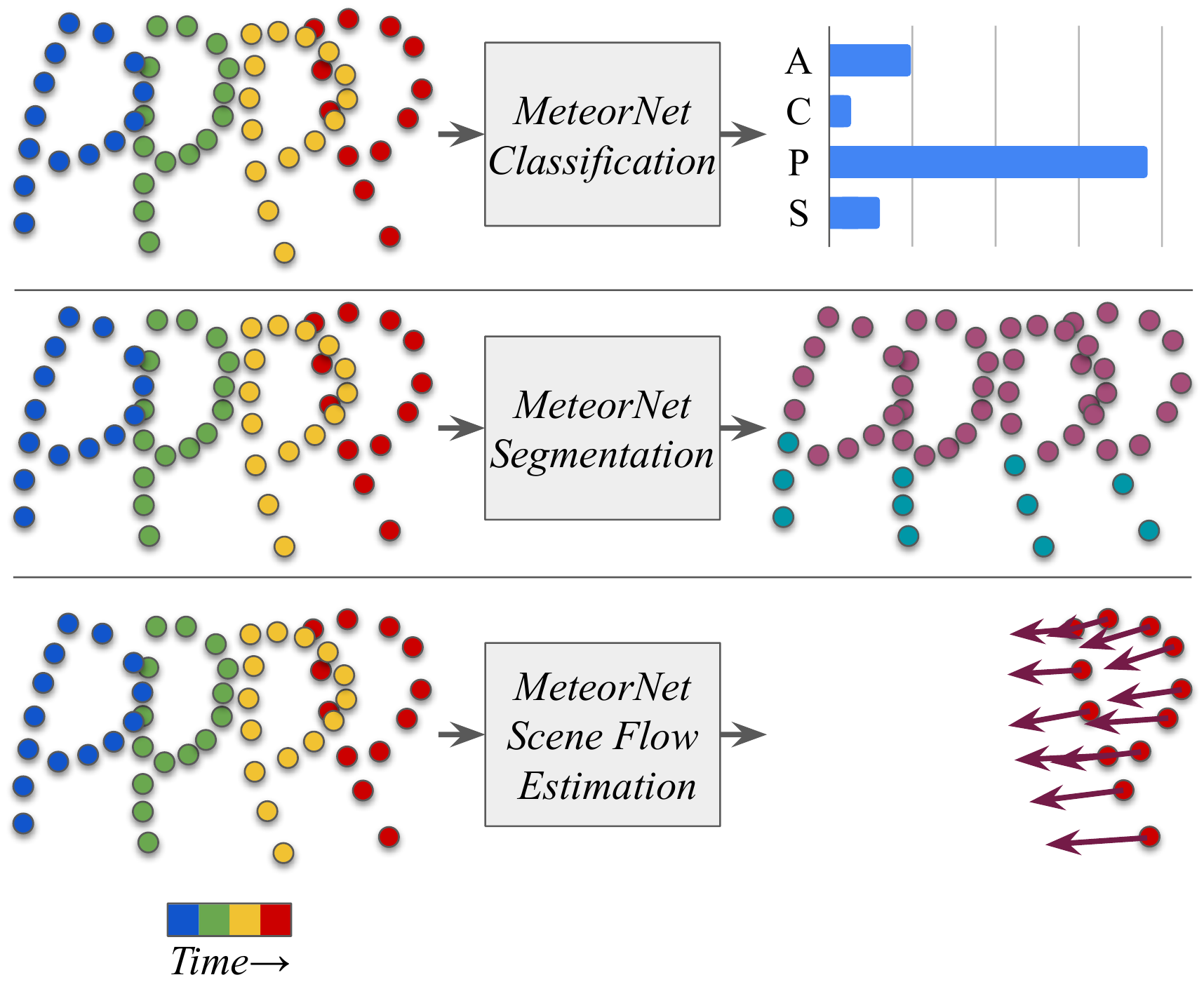}
\caption{ \textbf{MeteorNet applications.} Our model directly consumes a dynamic sequence of raw point clouds, and learns both global and local point features for various applications including classification, semantic segmentation and scene flow estimation. 
}
\label{fig:teaser}
\vspace{-1ex}
\end{figure}

In this work, we present a novel method for learning representations for dynamic 3D point cloud sequences. 
The key to our approach is a novel neural network module named the \emph{Meteor} module. This module takes in point cloud sequences and aggregates information in spatiotemporal neighborhoods to learn features for each point. The module can be stacked on top of each other, where per-point features from the previous module are input to the next module. The stacked modules hierarchically aggregate information from larger neighborhoods. Inspired by \cite{PointNet,PointNet++,Dynamic:Graph:CNN}, the aggregation process is implemented by applying the same multi-layer perceptrons (MLPs) to each point in the neighborhood and max pooling afterwards. 

We propose two methods for determining the spatial-temporal neighborhoods to address the motion range of an object: direct grouping and chained-flow grouping. The former method directly increases the grouping radius over time. The latter method tracks object motions and uses off-line estimated scene flow to construct efficient and effective neighborhoods. We conduct intensive experiments to compare these two methods. 

As visualized in Figure~\ref{fig:teaser}, learned features from Meteor modules can be used for downstream tasks such as classification, segmentation or scene flow estimation. We name the resulting deep neural network \emph{MeteorNet}. 
For semantic segmentation, MeteorNet achieved state-of-the-art results on the dataset Synthia \cite{Synthia}. It also outperforms previous methods on a semantic segmentation dataset derived from KITTI \cite{menze2015object}. MeteorNet specifically showed its advantage for recognizing movable objects on these two datasets. For scene flow estimation, MeteorNet beats previous baselines and achieves leading performance on FlyingThings3D \cite{Flyingthings3D:Driving} and the KITTI scene flow dataset \cite{KITTI:Scene:flow}. It also achieves leading performance on the action recognition dataset MSRAction3D \cite{MSR:Action:3D}.
To the best of our knowledge, this is the very first work on deep learning for dynamic \emph{raw} point cloud sequences. We expect that our MeteorNet can benefit research and applications in autonomous driving, robotic manipulation and related domains.

\section{Related Work}

\textbf{Deep learning for RGB videos } 
Existing approaches towards deep learning on videos can be categorized by how the temporal relationship between frames is modelled. The first family of approaches extracts a global feature for each video frame with a shared CNN and uses recurrent neural nets to model temporal relations \cite{Long-Term:RCNN,Beyond:Short:Snippets}. The second family of approaches learns temporal relations from offline-estimated optical flow \cite{Two-stream:Fusion}
or optical flow trajectories \cite{Two-stream:CNN} with a separate branch of the network besides the RGB branch. 
 The third family of approaches uses 3D CNNs and learns temporal relations implicitly   \cite{C3D,I3D,Karpathy:Video:CNN,ARTNet,ECO}. 
The fourth family of approaches uses non-local operations \cite{NLNet} or correspondence proposals \cite{CPNet} to learn long-range dependencies.
 Our work is a deep learning method for 3D videos and is inspired by the above methods. 

\textbf{Grid-based 3D deep learning }
Different representations for 3D geometry have been discussed in the literature \cite{3D:Deep:Learning:Survey}. A 3D occupancy grid is one of the most popular representations. Previous works have explored 3D convolution \cite{3d:shapenets, VoxNet} or sparse 3D convolution \cite{SECOND} for various 3D recognition tasks.
Recent works on deep learning for 3D sequences used a 4D occupancy grid representation by adding an additional time dimension. Fast-and-Furious \cite{FaF} proposed to view the vertical dimension as feature channels and apply 3D convolutions on the remaining three dimensions. MinkowskiNet \cite{MinkNet} explicitly used sparse 4D convolution on a 4D occupancy grid. Instead of quantizing the raw point clouds into an occupancy grid, our method directly processes point clouds. 

\textbf{Deep learning on 3D point clouds } 
Another popular representation of 3D geometry is 3D point clouds. Two major categories of methods have been explored.
The first category is based on PointNet \cite{PointNet}. The core idea is a symmetric function constructed with shared-weight deep neural networks applied to every point followed by an element-wise max pooling.
Follow-up work is PointNet++ \cite{PointNet++} which extracts local features of local point sets within a neighborhood in Euclidean space and hierarchically aggregates features. Dynamic graph CNN \cite{Dynamic:Graph:CNN} proposed a similar idea. The difference is that the neural network processes point pairs instead of individual points. FlowNet3D \cite{FlowNet3D} lets the shared neural network take mixed types of modalities, i.e. geometric features and displacement, as inputs to learn scene flow between two point clouds. 

The second category of methods combines the grid and point representation. 
VoxelNet \cite{VoxelNet} divides the space into voxels, uses local PointNets within each voxel and applies 3D convolution to voxel grids. SPLATNet \cite{splatnet} interpolates the point values to grids and applies 3D convolution before interpolating back to the original point cloud.
Our work lies in the first category and focuses on learning representations for point cloud sequences.

\begin{figure*}[h]
\centering
\includegraphics[width=\textwidth]{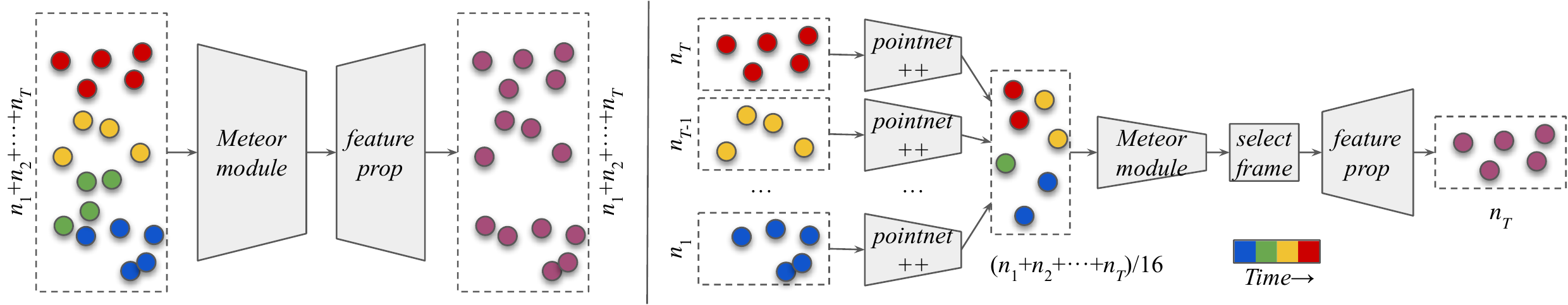}
\caption{Architecture design choices for MeteorNet. Left: Early fusion with per-point output for all frames. Right: Late fusion with per-point output for the last frame.}
\label{fig:overall:arch}
\end{figure*}

\section{Deep Learning on 3D Point Cloud Sequences}
\label{sec:method}

Our proposed method 
addresses the following three properties of point cloud sequences:

1. \textbf{Unordered intra-frame.} Points within the same frame should be treated as an unordered set. Change of feeding order of points within frames should not change the output of the deep learning model.

2. \textbf{Ordered inter-frame.} Points in different frames should be distinguished by their time stamps. Changing the time stamp of a point means moving the point to a different frame and should change the resulting feature vector.

3. \textbf{Spatiotemporal metric space.} Neighboring points form a meaningful local structure. For point cloud sequences, points that are close spatially {\em and\/} temporally should be considered neighbors. Therefore, the metric space that defines the neighborhood should include both the spatial and temporal domains.

In this section, we first briefly review point cloud deep learning techniques. Then we describe the Meteor module, the core module of our network, which serves the above three properties. We also explain the overall architecture design choices for various downstream applications. 
Finally, we present the theoretical foundation of universal approximation for our architecture.

\subsection{Review of PointNet}
\label{sec:reviewPCDL}
Our method is inspired by the seminal work of PointNet \cite{PointNet}, a neural network architecture for deep learning on a single point cloud.  It has been proven in \cite{PointNet} that given a set of point clouds $\mathcal{X} \subseteq \{ \{\mathbf{x}_1, \mathbf{x}_2, \ldots, \mathbf{x}_n\} \mid n\in \mathbb{Z}^+, \mathbf{x}_i \in [0, 1]^m \}$ and any continuous set function $f:\mathcal{X} \rightarrow \mathbb{R}^c$ w.r.t the Hausdorff distance, there exists a set function in the form of
$$
g(S) = \gamma \circ \underset{\mathbf{x}_i \in S}{MAX} \{ \eta(\mathbf{x}_i)\}    
$$
that can approximate $f$ on $\mathcal{X}$ arbitrarily closely. $\eta: \mathbb{R}^m \rightarrow \mathbb{R}^r$ and $\gamma: \mathbb{R}^r \rightarrow \mathbb{R}^c$ are two continuous functions and $MAX$ is the element-wise maximum operation. In practice, $\eta$ and $\gamma$ are instantiated to be MLPs. The follow-up work of PointNet++ \cite{PointNet++} extracts features in local point sets within a neighborhood and hierarchically aggregates features. For a point $\mathbf{x}_i \in S$, its learned feature is
$$
g'(\mathbf{x}_i) = \gamma \circ \underset{\{\mathbf{x}_j \mid \mathbf{x}_j \in S, \mathbf{x}_j \in \mathcal{N}_0(\mathbf{x}_i) \}}{MAX} \{ \eta(\mathbf{x}_j)\}
$$
where the neighborhood $\mathcal{N}_0(\mathbf{x}_i)$ can be decided by a fixed radius $r$ or the $k$ nearest neighbors of $\mathbf{x}_i$.

\subsection{Meteor Module}
\label{sec:meteor:module}

A point cloud sequence of length $T$ is defined as a $T$-tuple of 3D point clouds $S=(S_1, S_2, \ldots, S_T)$. Each element of this tuple is a 3D point set $S_t = \{p_i^{(t)} \mid i = 1,2,\ldots, n_t\}$, where the point $p_i^{(t)}$ is represented by  its Euclidean coordinates $\mathbf{x}_i^{(t)} \in \mathbb{R}^3$ and a feature vector $f_i^{(t)} \in \mathbb{R}^c$. 
The feature vector can be attributes obtained from sensors, e.g. color, or output from the previous Meteor module.
Our proposed Meteor module consumes $S$ as input and produces an updated feature vector  $h(p_i^{(t)})$ for every point $p_i^{(t)}$ in $S$. 
The first step of Meteor module is to find neighboring points of $p_i^{(t)}$ in the same or nearby frames to form a local spatiotemporal neighborhood $\mathcal{N}(p_i^{(t)})$. 

Given $\mathcal{N}$, we introduce two instantiations of $h$ for Meteor module.
The first instantiation is for applications where the correspondence across frame is important (e.g. scene flow estimation). 
For each ($p_j^{(t')}, p_i^{(t)})$ pair,  we pass the feature vectors of two points
and difference of their spatiotemporal positions 
into to an MLP with shared weights $\zeta$, followed by an element-wise max pooling. 
The updated feature of $p_i^{(t)}$ is then obtained by 
$$
h(p_i^{(t)}) = \underset{p_j^{(t')}\in \mathcal{N}(p_i^{(t)})}{MAX} \{ \zeta(f_j^{(t')}, f_i^{(t)}, \mathbf{x}_j^{(t')} - \mathbf{x}_i^{(t)}, t'-t) \}
$$
The second instantiation is for applications where  the correspondence between points across frames
is not important (e.g. semantic segmentation), we pass the feature vector of $p_j^{(t')}$ and difference of the spatiotemporal positions between  $p_j^{(t')}$ and  $p_i^{(t)}$ to $\zeta$ followed by a max pooling layer 
$$
h(p_i^{(t)}) = \underset{p_j^{(t')}\in \mathcal{N}(p_i^{(t)})}{MAX} \{ \zeta(f_j^{(t')}, \mathbf{x}_j^{(t')} - \mathbf{x}_i^{(t)}, t'-t) \}
$$
In terms of the spatiotemporal interaction range of a point  $\mathcal{N}$, we introduce two types of point grouping methods for deciding $\mathcal{N}$, which is crucial for the Meteor module:
\emph{direct grouping} and \emph{chained-flow grouping}.

\begin{figure}[h]
\centering
\small
\begin{subfigure}[b]{0.185\textwidth}
                \includegraphics[width=\linewidth]{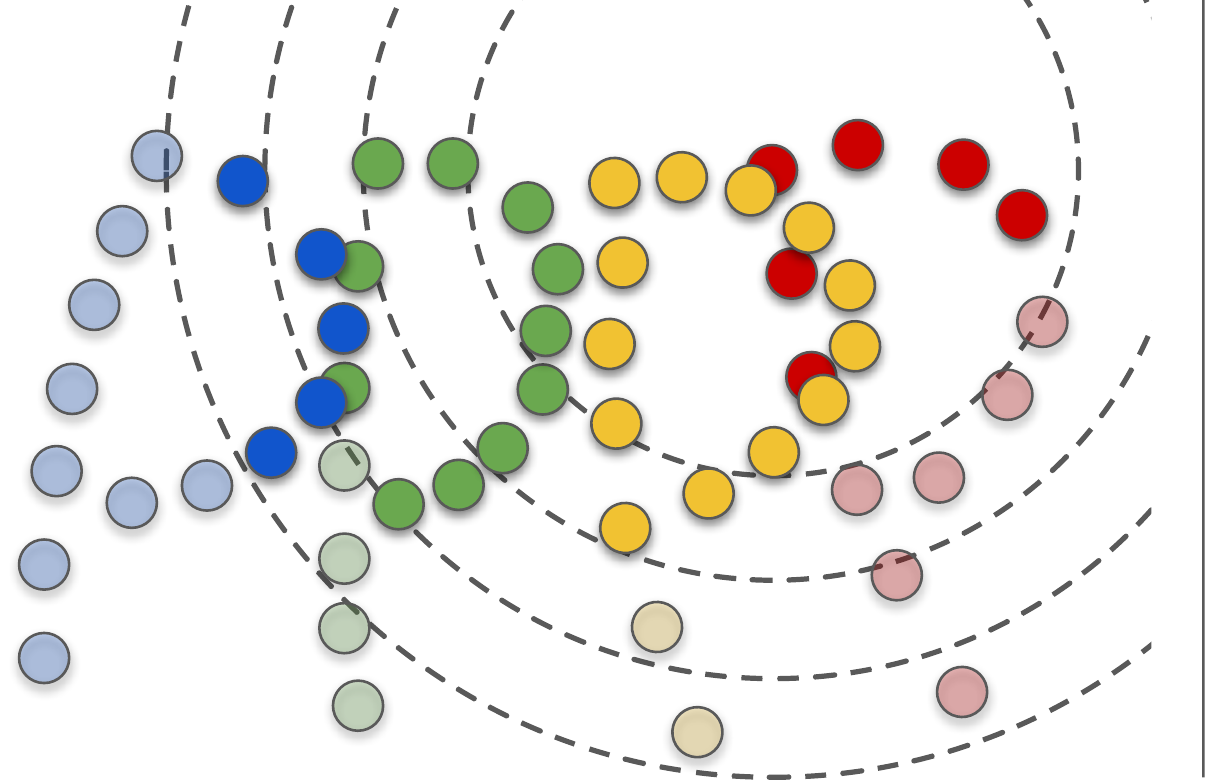}
                \caption{}
                \label{fig:direct_grouping}
        \end{subfigure}
\begin{subfigure}[b]{0.285\textwidth}
                \includegraphics[width=\linewidth]{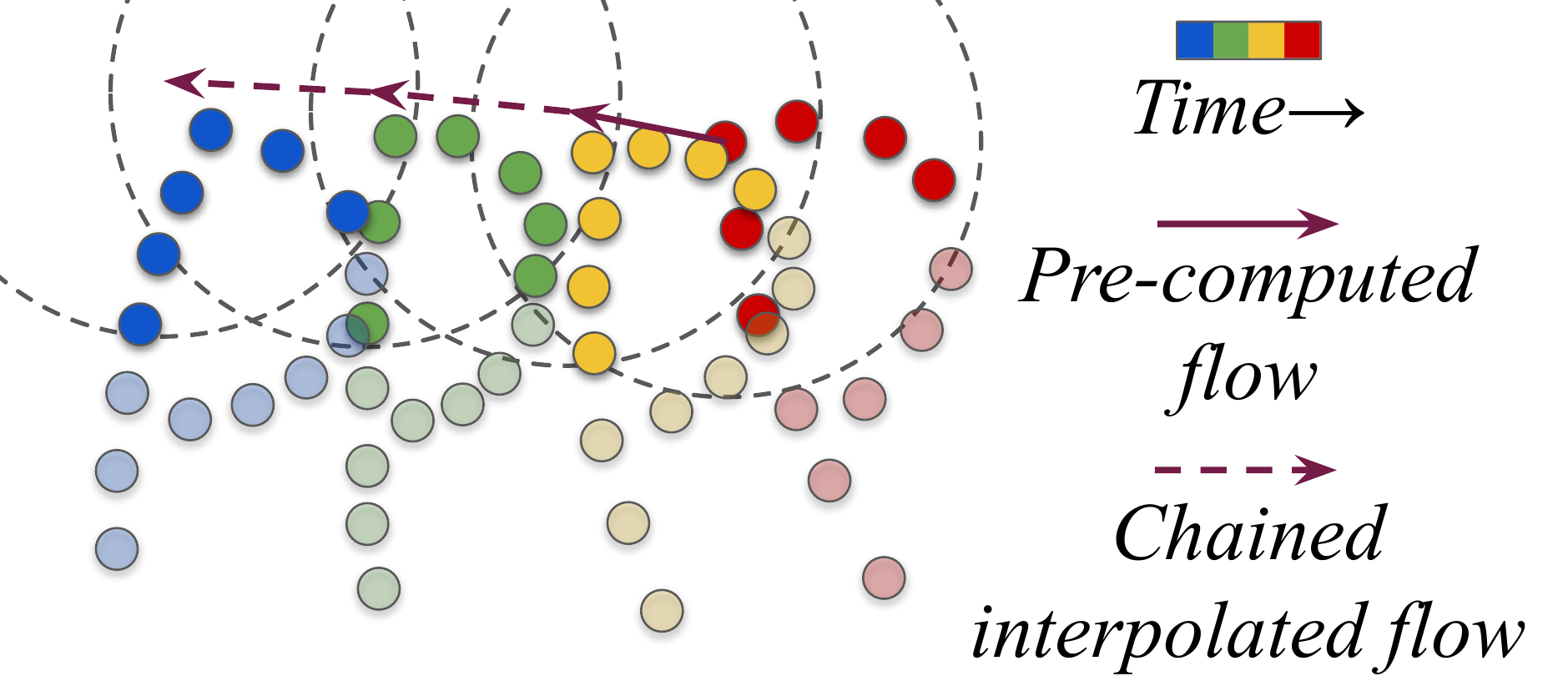}
                \caption{}
                \label{fig:chained_grouping}
        \end{subfigure}
\vspace{-5ex}
\caption{Two types of grouping methods: (a) direct grouping; (b) chained-flow grouping. 
}
\label{fig:grouping}
\vspace{-1ex}
\end{figure}

\textbf{Direct grouping.} Our intuition is that, the maximum distance an object can travel increases as time increases. Thus we directly increase grouping radius in 3D spatial space as $|t - t'|$ increases, to cover motion range across time. Formally, the neighborhood is decided by$$
\mathcal{N}_d (p_i^{(t)}; r) = \{p_j^{(t')} \mid ||\mathbf{x}_j^{(t')} - \mathbf{x}_i^{(t)}|| < r(|t'-t|) \}
$$
where $r$ is a monotonically increasing function. Note that $r(0) > 0$ so that points within the same frame can also be grouped. Direct grouping is illustrated in the Figure \ref{fig:grouping}(\subref{fig:direct_grouping}).

\textbf{Chained-flow grouping.}
In real-world dynamic point cloud sequences, the object represented by points in one frame usually has its corresponding points spatially close in neighboring frames. The motion trajectory of the object that is established through these correspondences is crucial for temporal understanding. 
In dynamic point cloud sequences, the interaction between a point and the other points in its spatiotemporal neighborhood should follow the direction of its motion. Such motion can be described by \emph{scene flow} ---  dense 3D motion field \cite{sceneflow}. 

In practice, for all $t$, we first estimate the backward scene flow $ \mathfrak{f}_i^{(t,t-1)} \in \mathbb{R}^3$ from frames $t$ to $t-1$
$$
\{ \mathfrak{f}_i^{(t,t-1)} \}_i = \mathcal{F}_0( \{p_i^{(t)} \},  \{p_j^{(t-1)} \})
$$
where $\mathcal{F}_0$ is a scene flow estimator between point clouds $\{p_i^{(t)} \}$ and $\{p_j^{(t-1)}\}$, e.g. FlowNet3D \cite{FlowNet3D}. 
Then ${\mathbf{x}_i'}^{(t-1)} = \mathbf{x}_i^{(t)} + \mathfrak{f}_i^{(t,t-1)} $  is the estimated position of the virtual corresponding point of $p_i^{(t)}$ in frame $t-1$. 

To estimate the corresponding position 
of $p_i^{(t)}$ in frame $t-2$, we first interpolate the scene flow estimation results $\{ \mathfrak{f}_j^{(t-1,t-2)} \}_j$ at the position of virtual point ${p_i'}^{(t-1)}$. 
Among choices for interpolation, we use the simple inverse distance weighted average of $k$ nearest neighbors
$$
{\mathfrak{f}_i'}^{(t-1,t-2)} = \frac{ \sum_{j=1}^k w(\mathbf{x}_j^{(t-1)}, {\mathbf{x}_i'}^{(t-1)})  \mathfrak{f}_j^{(t-1,t-2)}}{\sum_{j=1}^k w(\mathbf{x}_j^{(t-1)}, {\mathbf{x}_i'}^{(t-1)}) }
$$
where $w(\mathbf{x}_1, \mathbf{x}_2) = \frac{1}{d(\mathbf{x}_1, \mathbf{x}_2)^p}$ is the interpolation weight. We use $p=2, k=2$ by default. Then we chain the flow to estimate the corresponding position of $ p_i^{(t)}$ in frame $t-2$: ${\mathbf{x}_i'}^{(t-2)} = \mathbf{x}_i^{(t)} + \mathfrak{f}_i^{(t,t-1)} + {\mathfrak{f}_i'}^{(t-1,t-2)}$. 
Its position in frames beyond $t-2$ can be further interpolated and chained by repeating the above process. The pre-computed flow vectors $\mathfrak{f}$ and chained interpolated flow vectors $\mathfrak{f}'$ are illustrated in Figure \ref{fig:grouping}(\subref{fig:chained_grouping}).
Then the neighborhood for chained-flow grouping is determined by
$$
\mathcal{N}_c (p_i^{(t)}; r) = \{p_j^{(t')} \mid ||\mathbf{x}_j^{(t')} - {\mathbf{x}_i'}^{(t')}|| < r\}
$$
where $r$ is a constant value by default. One can also use a monotonically increasing function of $|t-t'|$ for  $r$ to compensate for scene flow estimation error.

Compared to direct grouping, chained-flow grouping focuses on tracking the motion trajectories and correspondences of each point so that a smaller grouping radius can be applied. It can also be more computationally efficient.

\subsection{Overall Architecture and Applications}

The feature vectors output by Meteor module can be further processed to obtain a quantity 
for the entire sequence, such as class scores (e.g. for classification); or propagated back to each point to obtain a per-point quantity, such as the class scores for all points (e.g. for semantic segmentation); or a per-point quantity for the points in a particular frame (e.g. scene flow estimation). We name the overall architecture for classification, semantic segmentation and scene flow estimation as MeteorNet-cls, MeteorNet-seg and MeteorNet-flow respectively.

There are generally two types of design choices for including Meteor modules in the architecture: Early fusion and Late fusion. 
 
\textbf{Early fusion } We apply Meteor module at the first layer so that the points from different frames are mixed from the beginning, as illustrated in left part of Figure \ref{fig:overall:arch}. In the following experiment section, MeteorNet-cls and MeteorNet-seg used early fusion.

\textbf{Late fusion } We apply a few layers of feature learning (e.g. PointNet++) individually to points in each frame before mixing them in the Meteor module, as illustrated in right part of Figure \ref{fig:overall:arch}. It allows the model to capture higher level
semantic features. In the follwing experiment section, MeteorNet-flow used late fusion.

\subsection{Theoretical Foundation}
We provide a theoretical foundation for our MeteorNet by showing the universal approximation ability of Meteor module to continuous functions on point cloud sequences.

Suppose $\forall t, \mathcal{X}_t = \{S_t \mid S_t\subseteq [0, 1]^m, |S_t| = n, n\in\mathbb{Z}^+ \}$ is the set of $m$-dimensional point clouds inside an $m$-dimensional unit cube at time $t \in \mathbb{Z}$. We define single-frame Hausdorff distance $d_H(S_i,S_j)$ for $S_i\in\mathcal{X}_i$ and $S_j\in\mathcal{X}_j$. $\mathcal{X}=\mathcal{X}_1 \times \mathcal{X}_2 \times \ldots \times \mathcal{X}_T$ is the set of point cloud sequences of length $T$. Suppose $f: \mathcal{X} \rightarrow \mathbb{R}$ is a continuous function on $\mathcal{X}$ w.r.t $d_{seq}(\cdot,\cdot)$, i.e. $\forall \epsilon > 0$, $\exists\delta>0$, for any $S, S^\prime \in \mathcal{X}$, if $d_{seq}(S, S^\prime) < \delta$, $|f(S)-f(S^\prime)| < \epsilon$. Here, we define the distance of point cloud sequences $d_{seq}(\cdot, \cdot)$ as the maximum per-frame Hausdorff distance among all respective frame pairs, i.e. $d_{seq}(S,S^\prime)=\max_t\{d_H(S_t, S_t^\prime)\}$. 
Our theorem says that $f$ can be approximated arbitrarily closely by a large-enough neural network and a max pooling layer with enough neurons.

\begin{theorem}
Suppose  $f: \mathcal{X}_1 \times \mathcal{X}_2 \times \ldots \times \mathcal{X}_T \rightarrow \mathbb{R}$ is a continuous function w.r.t  $d_{seq}(\cdot, \cdot)$. $\forall \epsilon > 0$, $\exists$ a continuous function $\zeta(\cdot, \cdot)$ and a continuous function $\gamma$,
such that for any $S=(S_1, S_2, \ldots, S_T) \in \mathcal{X}_1 \times \mathcal{X}_2 \times \ldots \times \mathcal{X}_T$, 
$$
\left|f(S) - \gamma \circ \Big(\underset{\mathbf{x}_i^{(t)}\in S_t, t\in \{1,2,\ldots,T\}}{MAX}\{\zeta(\mathbf{x}_i^{(t)}, t)\} \Big) \right| < \epsilon
$$
where $MAX$ is a vector max operator that takes a set of vectors as input and returns a new vector of the element-wise maximum.
\end{theorem}

The proof of this theorem is in the supplementary material. The key idea is to prove the homeomorphism between a point cloud sequence and a single-frame point cloud by designing a continuous bijection between them so that the conclusions for a single-frame point cloud can be used. We explain the two implications of the theorem. First, for a point cloud sequence of length $T$, function $\zeta$ essentially processes $T\times$ more points identically and independently than for a single-frame point cloud.
Thus compared to single-frame PointNet \cite{PointNet}, it may need a larger network as well as a larger bottleneck dimension for $MAX$ to maintain the expressive power for a point cloud sequence. We will show this in later experiments. Second, simply adding an additional time stamp $t$ as input to $\zeta$, the network is able to distinguish the points from different frames but still treats the points from the same frame as an unordered set.

\section{Grids versus Point Clouds: A Toy Example}
We constructed a toy dataset of point cloud sequences where grid-based methods fail in learning a simple classification task. Through this simple dataset, we show the drawbacks of these grid-based methods and the advantage of our architecture which works on raw point clouds.

\begin{figure}[h]
\centering
\includegraphics[width=0.4\textwidth]{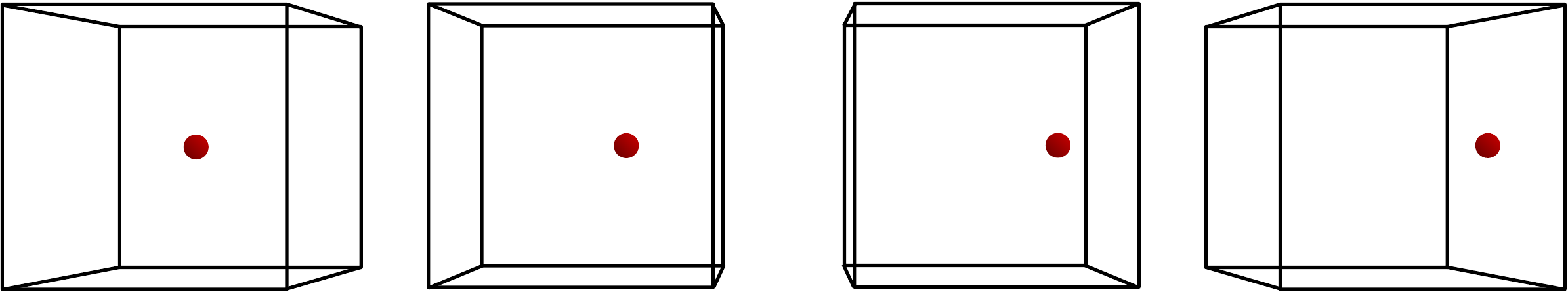}
\caption{A ``fast'' example in our dataset.}
\label{fig:toy:example}
\vspace{-2ex}
\end{figure}

The dataset consists of sequences of a randomly positioned particle moving inside a $100\times100\times100$ cube. The moving direction is randomly chosen from 6 possible directions, all parallel to one of the edges of the cube.
Each sequence has four frames. There are four categories for the motion speed of a particle: ``static'', ``slow'', ``medium'' and ``fast''. The moving distance of the particle at each step is zero for the ``static'' category and randomly chosen between $[0.09, 0.11]$, $[0.9, 1.1]$ and $[9, 11]$, respectively, for the other three categories. The dataset has 2000 training and 200 validation sequences, each with an equal number of sequences per label. Figure \ref{fig:toy:example} illustrates an example data point in our dataset with the ``fast'' label.

We used a 
toy version of two recent grid-based deep architectures for dynamic 3D scenes, FaF \cite{FaF} and MinkNet \cite{MinkNet}, as well as our MeteorNet. We only allow three layers of neurons within the three network architectures and convolution kernel sizes of no larger than 3. The architecture details are listed in Table \ref{tab:toy}. The particle is represented by grid occupancy for grid-based methods, and by a 3D point for MeteorNet. 
The experiment settings are designed to simulate situations where stacking convolution layers to increase expressive power is insufficient or inefficient. 
No data augmentation is used. The training and validation results are listed in Table \ref{tab:toy}. Our model can 
perfectly learn the toy dataset, while previous grid-based methods cannot avoid significant errors regardless of grid size chosen.

\begin{table}[h]
\small
\centering
\setlength{\tabcolsep}{1.pt}
\begin{tabular}{l|c|c|c|c}
\hline
\multicolumn{2}{l|}{} & FaF & MinkNet & \textbf{MeteorNet}  \\ 
\multicolumn{2}{l|}{} & \cite{FaF}  &  \cite{MinkNet}  & \textbf{(ours)} \\ \hline
\multirow{3}{*}{arch} & layer 1 & conv16, 3$\times$3$\times$3 & conv16, 3$\times$3$\times$3$\times$3  & \multirow{2}{*}{mlp [16,16]} \\ \cline{2-4} 
& layer 2 & conv16, 3$\times$3$\times$3 & conv16, 3$\times$3$\times$3$\times$3 & \\ \cline{2-5}
& layer 3 & \multicolumn{3}{c}{max pool, fc} \\ \specialrule{.1em}{.05em}{.05em}
\multirow{6}{*}{\begin{tabular}[c]{@{}l@{}}grid \\ size\end{tabular}} & 0.05 & N/A, N/A & 76.41, 75.00 & \multirow{6}{*}{\textbf{100.00}, \textbf{100.00}} \\ 
& 0.10 & N/A, N/A & 76.16, 75.50 & \\
& 0.50 & 91.05, 79.00 & 89.31, 87.50 & \\
& 1.00 & 84.20, 78.50 & 79.73, 82.00 & \\
& 5.00 & 61.15, 63.50 & 59.87, 61.00 & \\
& 10.0 & 59.65, 60.00 & 51.41, 53.00 & \\ \hline
\end{tabular}
\caption{Toy experiment settings and results. Accuracy results (\%) are shown in ``\textit{train}, \textit{val}''. N/A denotes resource (e.g.memory) not enough.}
\vspace{-1ex}
\label{tab:toy} \end{table}

We provide an explanation as follows. If the grid size is chosen to be comparable to the ``slow'' step size, ``static'' and ``slow'' particles can be correctly classified. However, both ``medium'' or ``fast'' particles require huge convolution receptive fields to cover the moving distance. Thus, the toy grid-based networks failed to classify them.
If the grid size is chosen to be comparable to the ``fast'' (or ``medium'') step size, ``fast'' (or ``medium'') particles can be correctly classified. However, both ``slow'' and ``static'' particles mostly yield similar stationary grid occupancy. Thus, the toy grid-based networks failed to classify them.
Through this experiment, we show 
the advantage
of our MeteorNet: it is grid size agnostic and can accurately learn both long- and short-range motion at the same time.

\section{Experiments}
The design of our MeteorNet is motivated by two hypotheses. First, deep architectures that process raw point clouds suffer less from quantization error than grid-based methods. Therefore, they achieve a better performance in a variety of recognition tasks. Second, many recognition tasks benefit from longer input sequences of 3D data. To show this, we apply MeteorNet to three 3D recognition tasks: action recognition, semantic segmentation and scene flow estimation, which are typical classification and regression tasks. We compare the performance to a variety of baselines including grid-based and single-frame methods. We also visualize some example results.

\subsection{Classification}
We first conducted an experiment on point cloud sequence classification. We use the MSRAction3D dataset \cite{MSR:Action:3D}. It consists of 567 Kinect depth map sequences of 10 different people performing 20 categories of actions. There are 23,797 frames in total. We reconstructed point cloud sequences from the depth maps. We use the same train/test split as previous work \cite{actionlet}. The  classification results are listed in Table \ref{tab:msr:cls}.
The baselines are descriptor-based methods on depth maps \cite{Stop:descriptor,klaser:descriptor} and skeletons \cite{actionlet}, and PointNet++ \cite{PointNet++}. MeteorNet-cls significantly outperforms all baselines on this dataset. We also show that action recognition accuracy of MeteorNet-cls benefits from longer point cloud sequences.

\begin{table}[t]
\small
\setlength{\tabcolsep}{3.pt}
\centering
\begin{tabular}{c|c|c|c}
\hline
Method & Input & \# of Frames & Accuracy \\  \hline
Vieira et al. \cite{Stop:descriptor} & depth & 20 & 78.20 \\
Kl\"{a}ser et al. \cite{klaser:descriptor} & depth & 18 & 81.43 \\
Actionlet \cite{actionlet} & groundtruth skeleton & full & 88.21 \\ \hline
PointNet++ \cite{PointNet++} & point & 1 & 61.61  \\  \hline
\multirow{5}{*}{\textbf{MeteorNet-cls}} &  & 4 & 78.11  \\ 
&  & 8 & 81.14  \\ 
& point & 12 & 86.53  \\ 
&  & 16 & 88.21  \\
&  & 24 & \textbf{88.50}  \\ \hline
\end{tabular}
\vspace{-1ex}
\caption{Classification accuracy on MSRAction3D (\%).
}
\label{tab:msr:cls}
\vspace{-1ex}
\end{table}

\begin{table*}[h]
\small
\setlength{\tabcolsep}{1.1pt}
\centering
\begin{tabular}{c|c|c|cccccccccccc|cc}
\hline
\multirow{2}{*}{Method} & Params & \# of & \multicolumn{12}{c|}{IoU} & \multirow{2}{*}{mIoU} & \multirow{2}{*}{mAcc} \\
& (M) & Frames & Bldng & Road & Sdwlk & Fence & Vegitn & Pole  & Car & T.sign & Pdstr & Bicyc & Lane & T.light &  \\ \hline
3D MinkNet14 \cite{MinkNet} & 19.31 & 1 & 89.39 & 97.68 & 69.43 & 86.52 & 98.11 & 97.26 & 93.50 & 79.45 & 92.27 & 0.00 & 44.61 & 66.69 & 76.24 & \textbf{89.31} \\
4D MinkNet14 \cite{MinkNet} & 23.72 & 3 & 90.13 & \textbf{98.26} & 73.47 & 87.19 & \textbf{99.10} & 97.50 & 94.01 & 79.04 & \textbf{92.62} & 0.00 & 50.01 & 68.14 & 77.46 & 88.01 \\ \specialrule{.1em}{.05em}{.05em}
PointNet++ \cite{PointNet++} & 0.88 & 1 & 96.88 & 97.72 & 86.20 & 92.75 & 97.12 & 97.09 & 90.85 & 66.87 & 78.64 & 0.00 & 72.93 & 75.17 & 79.35 & 85.43  \\ \hline
\textbf{MeteorNet-seg-\boldmath{$s$} (direct)} & 0.88 & 2 & 98.08 & 97.77 & 87.16 & 93.53 & 96.91 & 97.47 & 94.04 & 77.22 & 72.19 & 0.00 & \textbf{73.59} & 75.75 & 80.31 & 85.92  \\
\textbf{MeteorNet-seg-\boldmath{$m$} (direct)} & 1.36 & 2 & 97.65 & 97.83 & 90.03 & 94.06 & 97.41 & \textbf{97.79} & 94.15 & 82.01 & 79.14 & 0.00 & 72.59 & \textbf{77.92} & 81.72 & 86.57  \\
\textbf{MeteorNet-seg-\boldmath{$m$} (chain)} & 1.36 & 2& 98.22 & 97.79 & 90.98 & 93.18 & 98.31 & 97.45 & \textbf{94.30} & 76.35 & 81.05 & 0.00 & 74.09 & 75.92 & 81.47 & 86.42 \\
\textbf{MeteorNet-seg-\boldmath{$m$} (direct)} & 1.36 & 3 & \textbf{98.45} & 97.92 & \textbf{91.57} & \textbf{94.40} & 97.54 & 97.46 & 94.11 & 79.04 & 75.04 & 0.00 & 73.17 & 74.93 & 81.13 & 85.80  \\
\textbf{MeteorNet-seg-\boldmath{$l$} (direct)} & 1.78 & 3 & 98.10 & 97.72 & 88.65 & 94.00 & 97.98 & 97.65 & 93.83 & \textbf{84.07} & 80.90 & 0.00 & 71.14 & 77.60 & \textbf{81.80} & 86.78 \\
\hline 
\end{tabular}
\vspace{-0ex}
\caption{Semantic Segmentation results on the Synthia dataset.
Metrics are mean IoU and mean accuracy (\%).}
\label{tab:synthia:seg}
\vspace{-0ex}
\end{table*}

\subsection{Semantic Segmentation}
We conduct two semantic segmentation experiments. We first test MeteorNet-seg on large-scale synthetic dataset Synthia \cite{Synthia} to perform an ablation study and compare with a sparse 4D CNN baseline. Then we test MeteorNet-seg on real LiDAR scans from  KITTI dataset \cite{KITTI:raw}.

\textbf{Synthia dataset } It consists of six sequences of driving scenarios in nine different weather conditions. 
Each sequence consists of four stereo RGBD images 
from four viewpoints captured from the top of a moving car. 
We reconstruct 3D point clouds from RGB and 
depth images and create 3D point cloud sequences. The scene is cropped by a 50m$\times$50m$\times$50m bounding box centered at the car. We then use farthest point sampling to downsample the scene to 16,384 points per frame. 
We used the same train/validation/test split as \cite{MinkNet}: sequences 1-4 with weather conditions other than sunset, spring and fog are used as train set; sequence 5 with foggy weather are used as validation set; and sequence 6 with sunset and spring weather are used as test set. The train, validation and test set contain 19,888, 815 and 1,886 frames respectively. 

As an ablation study, we used MeteorNet-seg with various settings of model sizes and number of input frames.
Compared to MeteorNet-seg-$s$, MeteorNet-seg-$m$ has a larger bottleneck dimension at each max pooling layer. Compared to MeteorNet-seg-$m$, MeteorNet-seg-$l$ has the same max pooling dimensions but larger dimensions in non-bottleneck layers. Among the baselines, 3D and 4D MinkNet \cite{MinkNet} voxelize the space and use 3D spatial or 4D spatiotemporal sparse convolution to perform segmentation. 
The evaluation metrics are the per-class and overall mean IoU as well as overall segmentation accuracy. The results are listed in Table \ref{tab:synthia:seg}. 

\begin{figure}[t]
\centering
\small
\includegraphics[width=1\linewidth]{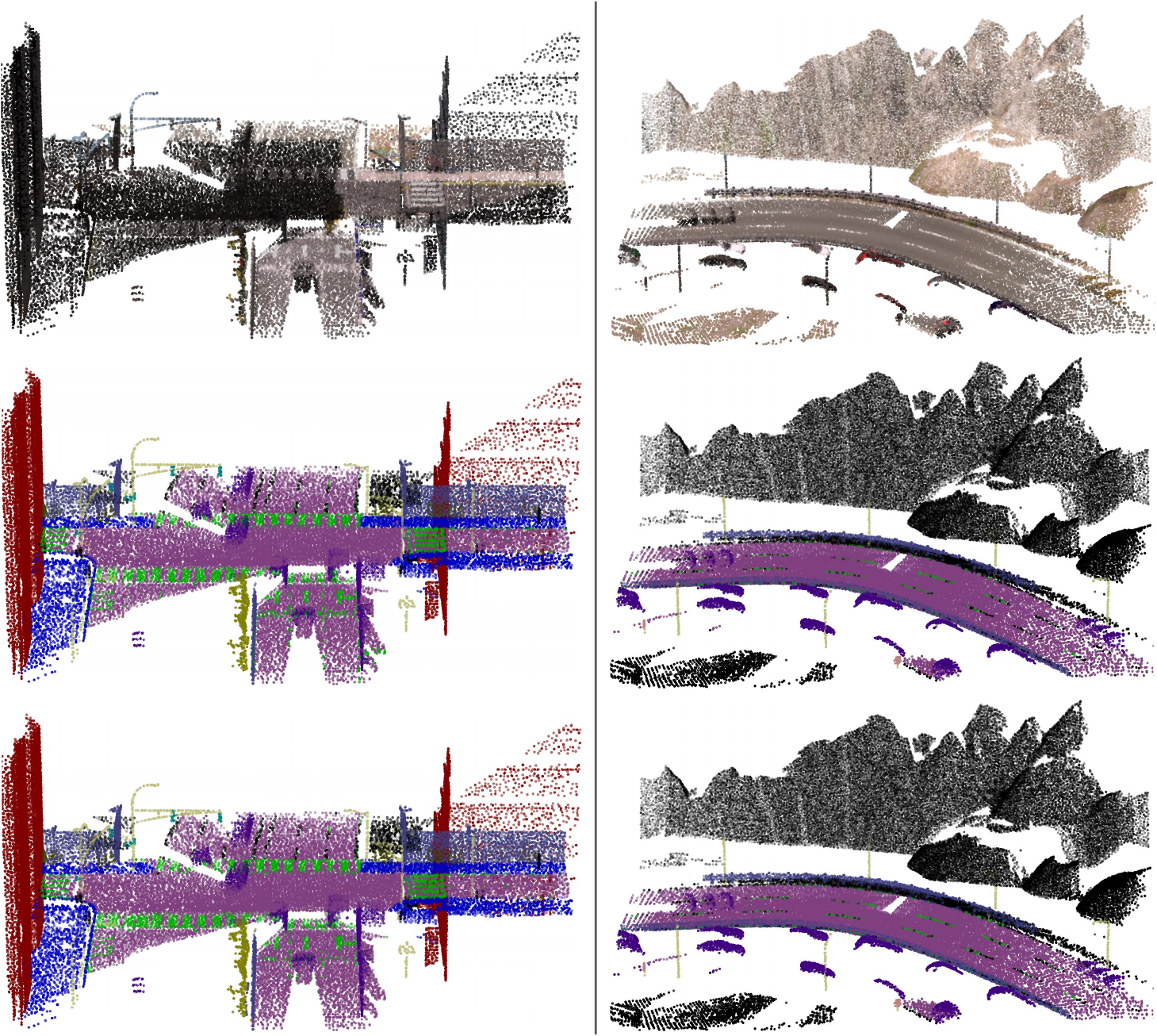}
\caption{ Visualization of two example results from the Synthia dataset. From the top: RGB input, ground truth, predictions.
}
\label{fig:viz:synthia}
\vspace{-1ex}
\end{figure}

Our multi-frame MeteorNet-seg outperforms the single-frame PointNet++. Furthermore, point-based methods outperform sparse-convolution-based methods for IoU by a significant margin in most categories as well as mean IoU.
We also found interesting results in the ablation study. First, increasing the max pooling bottleneck dimension is more effective than increasing the non-bottleneck layer dimension for multi-frame input.
Second, increasing the number of frames without increasing the model size can negatively impact the network performance.
Third, MeteorNet-seg with chained-flow grouping achieves the best performance on movable objects such as ``Car'' and ``Pedestrian'' among all point-based methods. This shows its advantage in detecting motion.
Figure \ref{fig:viz:synthia} visualizes two example results. Our model can accurately segment most objects. 

\textbf{KITTI dataset } We derived a semantic segmentation dataset from the KITTI object detection dataset \cite{kitti-3d-detection} by converting bounding box labels to per-point labels. 
Additional frames are from the corresponding LiDAR data in the KITTI raw set \cite{KITTI:raw}. 
We used GPS/IMU data to transform the point cloud into world coordinates to eliminate ego motion.
There are three semantic classes: ``Car'', ``Pedestrain/Cyclist'' and ``Background''. The ``Car'' class mostly consists of points of parked and static cars\footnote{We provide an approximate statistics by randomly sampling 200 point cloud sequences from all 7481 train/val sequences and count "Car" instances. There are totally 884 "Car" instances. Among them, 654 (73.9\%) are static cars, 230 (26.1\%) are moving cars.} while the ``Pedestrain/Cyclist'' class mostly consists of points of moving people. We used the default train/val split as in \cite{Frustum:PointNet}. The segmentation results are listed in Table \ref{tab:kitti:seg}.

\begin{table}[t]
\small
\setlength{\tabcolsep}{3.9pt}
\centering
\begin{tabular}{c|c|ccc|c}
\hline
\multirow{2}{*}{Method} & \# of & \multirow{2}{*}{Car} & Pdstr/ & \multirow{2}{*}{Bkgrd} & \multirow{2}{*}{mIoU} \\ 
& Frames & & Cyclst & \\ \hline
PointNet++ \cite{PointNet++} & 1 & 74.06 & 36.43 & 98.19 & 69.56  \\ \specialrule{.1em}{.05em}{.05em}
\multirow{2}{*}{\begin{tabular}[c]{@{}c@{}}\textbf{MeteorNet-seg} \\ \textbf{(direct)} \end{tabular}}  & 2 & \textbf{74.53}  & 42.19 & 98.24 & 71.65  \\ 
& 3 & 71.22  & \textbf{50.93} & 98.12 & \textbf{73.42} \\ \hline
\end{tabular}
\vspace{-0ex}
\caption{\textbf{Semantic Segmentation results on KITTI dataset.} Metrics are per-class and average IoU (\%).
}
\label{tab:kitti:seg}
\vspace{-1ex}
\end{table}

MeteorNet yields similar segmentation accuracy as PointNet++ on ``Car'' but significantly improves on ``Pedestrian/Cyclist'' with multiple frames as input. As expected, the accuracy on ``Pedestrian/Cyclist'' continues to increase as the number of input frames increases. This underlines the advantage of MeteorNet for learning object motion.

\subsection{Scene Flow Estimation}

Labelling dense scene flow in real point cloud data is very expensive. To the best of our knowledge, there does not exist any large-scale real-world point cloud dataset with per-point scene flow annotations.
A common approach is to train on a large-scale synthetic dataset and then to finetune on a smaller real dataset \cite{FlowNet3D}. To this end, we conducted two experiments:  we first train MeteorNet on the FlyingThings3D dataset \cite{Flyingthings3D:Driving}, then finetune MeteorNet on the KITTI scene flow dataset \cite{KITTI:Scene:flow}.

\textbf{FlyingThings3D dataset } This synthetic dataset consists of RGB and disparity image videos 
rendered from scenes of multiple randomly moving objects from ShapeNet \cite{ShapeNet}. It provides 8,955 training videos and 1,747 test videos where each video has 10 frames.  
We reconstructed 3D point clouds from 
disparity maps. Maps of optical flow and disparity change are provided for consecutive frames, from which 3D scene flow can be reconstructed.
This dataset is challenging because of large displacements and strong occlusions. 
We randomly sampled 20,000 4-frame sequences from training videos as our training set and 2,000 4-frame sequences from testing videos as our test set. We used random rotation for data augmentation. 

\begin{table}[h]
\small
\setlength{\tabcolsep}{3.pt}
\centering
\begin{tabular}{c|cc|cc|cc}
\hline
Method & Input & Frames & mean & std & acc & outlier \\ \hline
FlowNet-C \cite{dosovitskiy2015flownet} & depth & 2 & 0.473 & 0.275 & 10.75  & 11.60 \\
FlowNet-S \cite{dosovitskiy2015flownet} & depth & 3  & 0.437 & 0.281 & 22.25 & 10.62  \\ \hline
FlowNet3D \cite{FlowNet3D} &  points & 2 &  0.218 & 0.196  &   49.46    &  2.37   \\ \specialrule{.1em}{.05em}{.05em}
\multirow{2}{*}{\begin{tabular}[c]{@{}c@{}}\textbf{MeteorNet-flow} \\ \textbf{(direct)} \end{tabular}} & points & 3 & 0.219 & 0.187 & 47.44 & 2.30  \\ 
& points & 4 & 0.214 & 0.190 & \textbf{52.12} & 2.40  \\ \hline
\multirow{2}{*}{\begin{tabular}[c]{@{}c@{}}\textbf{MeteorNet-flow} \\ \textbf{(chained-flow)} \end{tabular}} & points & 3 & 0.215 & 0.194 & 49.63 & 2.44  \\ 
& points & 4 & \textbf{0.209} & \textbf{0.184} & 49.91 & \textbf{2.28} \\ \hline
\end{tabular}
\vspace{-0ex}
\caption{\textbf{Flow estimation results on the FlyingThings3D dataset.} Metrics are for the end-point-error (EPE) of scene flow: mean, standard deviation, accuracy (\%, ratio of estimations with EPE $<$0.1 or 10\%), and outlier ratio (\%, of estimations with EPE $>$1.0).
}
\label{tab:flyingthing3d}
\vspace{-1ex}
\end{table}

We evaluate 3D end-point-error (EPE) of scene flow, which is defined as the $L_2$ distance between the estimated flow vectors and the ground truth flow vectors. We report four aspects of EPE: mean, standard deviation, accuracy with threshold 10\% or 0.1, and outlier ratio with threshold 1.0, as our evaluation metrics. Among the baselines, FlowNet-C/FlowNet-S are convolutional architectures adapted from \cite{dosovitskiy2015flownet} that take two/three dense depth maps (converted to $xyz$ coordinate maps) as input to estimate per-pixel scene flow instead of optical flow as originally done in \cite{dosovitskiy2015flownet}. FlowNet3D \cite{FlowNet3D} is a PointNet-based architecture that takes two point clouds as input to estimate per-point scene flow. The results are listed in Table \ref{tab:flyingthing3d}. 

We see that convolution-based methods have a hard time  capturing accurate scene flow probably due to occlusion. 
Point based methods such as FlowNet3D can effectively capture the accurate motion in point clouds. MeteorNet-flow can further improve scene flow estimation with more frames as input. MeteorNet-flow with direct grouping shows better accuracy for small displacements while MeteorNet-flow with chained-flow grouping shows better overall performance. 

By using chained flow, points are grouped into a neighborhood which are located close to the proposed corresponding position as predicted by flow from past frames. Therefore, the model is provided more evidence to estimate the correct flow direction and magnitude. Furthermore, as the number of input frames increases for MeteorNet-flow, the performance gain is consistent.

\textbf{KITTI scene flow dataset } 
This dataset provides ground truth disparity maps and optical flow for 200 frame pairs \cite{KITTI:Scene:flow}. From this, we reconstructed 3D scene flow. 
Only 142 out of 200 pairs have corresponding raw LiDAR point cloud data and thus allow us to use preceding frames from the KITTI raw dataset \cite{KITTI:raw}.
We project the raw point clouds onto the groundtruth maps to obtain the groundtruth 3D scene flow.
We first train the models on the FlyingThings3D dataset until convergence and use first 100 pairs to finetune the models. Then we use the remaining 42 pairs for testing. 
We report two aspects of EPE of scene flow: mean and standard deviation, as our evaluation metrics. 
The baseline FlowNet3D is trained and finetuned the same way as MeteorNet-flow. The results are listed in Table \ref{tab:kitti:flow}. 

\begin{table}[t]
\small
\centering
\begin{tabular}{c|cc|cc}
\hline
Method   & Input & Frames & mean & std  \\ \hline
FlowNet3D \cite{FlowNet3D} & points & 2 & 0.287  & 0.250 \\ \specialrule{.1em}{.05em}{.05em}
\multirow{2}{*}{\begin{tabular}[c]{@{}c@{}}\textbf{MeteorNet-flow} \\ \textbf{(direct)} \end{tabular}}  & points & 3  & 0.282 & \textbf{0.204} \\ 
& points & 4  & 0.263 & 0.210 \\ \hline
\multirow{2}{*}{\begin{tabular}[c]{@{}c@{}}\textbf{MeteorNet-flow} \\ \textbf{(chained-flow)} \end{tabular}}  & points & 3  & 0.277 & 0.244   \\ 
& points & 4  & \textbf{0.251} & 0.227  \\ \hline
\end{tabular}
\vspace{-0ex}
\caption{\textbf{Flow estimation results on KITTI sceneflow dataset.} Metrics are the mean and standard deviation of the End-point-error (EPE) of scene flow.
}
\label{tab:kitti:flow}
\vspace{-1ex}
\end{table}

\begin{figure}[h]
\centering
\small
\includegraphics[width=1\linewidth]{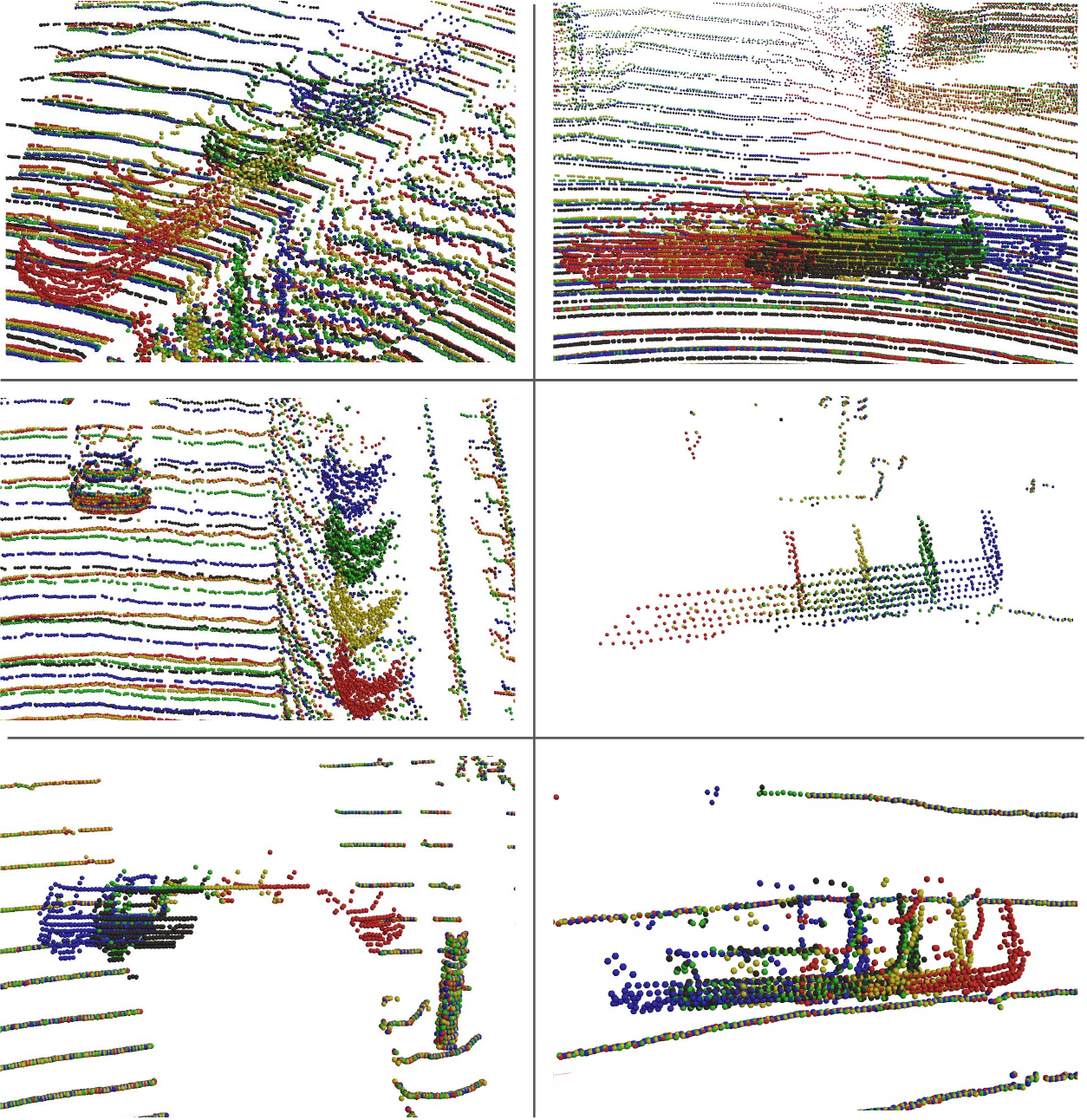}
\caption{{\bf Visualization of MeteorNet example results on the KITTI scene flow dataset.} Point are colored to indicate which frames they belong to: \textcolor{blue}{frames $t-3$}, \textcolor{viz_green}{frame $t-2$}, \textcolor{viz_yellow}{frame $t-1$}, \textcolor{viz_red}{frame $t$}. 
\textbf{Translated points} (frame $t-3$ + estimated scene flow) is in black. Green and black shapes are supposed to overlap for perfect estimation.
}
\label{fig:viz:kitti:flow}
\vspace{-0ex}
\end{figure}

Our MeteorNet-flow with three and four frames as input outperforms the baselines. As the number of frame increases, the performance gain is consistent. The version using chained-flow for grouping is better in mean error while the direct grouping version has lower standard deviation and is more robust. 
Our hypothesis is that when points are sparse, the initial flow estimation may have larger error which propagates through time and thereby affects the final prediction.
Figure \ref{fig:viz:kitti:flow} visualizes some examples of the resulting flow estimates. 

\section{Discussion}

\textbf{Relation to Other Architectures }
MeteorNet can be seen as a generalization of several previous architectures. 
When the input is one point cloud frame, MeteorNet can be reduced to either PointNet++\cite{PointNet++} or DGCNN \cite{Dynamic:Graph:CNN}, depending on the instantiation of the $h$ function in Subsection \ref{sec:meteor:module}. When the input has two frames of point clouds, MeteorNet can be reduced to FlowNet3D \cite{FlowNet3D}.

\textbf{Direct Grouping vs. Chained-flow Grouping }
In subsection \ref{sec:meteor:module}, we discussed the two grouping methods.
One possible advantage of chained-flow grouping compared to direct grouping is its computational efficiency. Intuitively for direct grouping, the spatial neighborhood radius $r$ should grow linearly with $t-t'$. Therefore, for a point cloud sequence with length $T$, the total number of points in the spatiotemporal neighborhood of a point is $\mathcal{O}(T^4)$. For chained-flow grouping, the spatial radius $r$ does not need to grow with time, thus the number of points in the spatiotemporal neighborhood is only $\mathcal{O}(T^2)$. This limitation of the direct grouping can be mitigated by limiting the temporal radius of neighborhoods while increasing the number of stacked Meteor modules. This is similar to using smaller convolution kernels while increasing the number of layers in convolutional neural networks.

A potential problem for chained-flow grouping is when point clouds are too sparse and the flow estimation is inaccurate. In this case, errors may accumulate during chaining, and the resulting spatiotemporal neighborhood may deviate from the true corresponding points across time. Studying the effect of initial scene flow error on final performance will be left as future work.

\section{Conclusion}
In this work, we proposed a novel deep neural network architecture MeteorNet that directly consumes dynamic 3D point cloud sequences. We show how this new architecture outperforms grid-based and single-frame methods on a variety of 3D recognition tasks including activity recognition, semantic segmentation and scene flow estimation. We also show the universal approximation ability of our network and provide visualizations of example results.

\section*{Acknowledgements}

Toyota Research Institute (``TRI'')  provided funds to assist the authors with their research but this article solely reflects the opinions and conclusions of its authors and not TRI or any other Toyota entity.

\newpage

{\small
\bibliographystyle{ieee_fullname}
\bibliography{egbib}
}

\appendix

\section{Overview}
In this document, we provide additional detail on MeteorNet as presented in the main paper. We present additional results on the accuracy of action recognition (Sec.~\ref{sec:msr:action:acc}) and the outlier ratio in scene flow estimation (Sec.~\ref{sec:scene:flow:outlier}). In Section \ref{sec:arch:detail}, we provide more details on the architectures used in various experiments. In Section \ref{sec:model:runtime}, we provide a runtime analysis for our model on the Synthia dataset. In Section \ref{sec:theorem}, we present the proof to our theorem. In Section \ref{sec:viz} we provide qualitative example results for various experiments.  Lastly, in Section \ref{sec:metaphor}, we give a brief rationale for the name of our neural network.

\section{MSRAction3D Per-class Accuracy }
\label{sec:msr:action:acc}

In the main paper, we showed that MeteorNet with multiple frames of point clouds as input outperforms various baselines. 
We obtained all possible clips of a certain length from a full-length point cloud sequence and computed the softmax classification scores on them individually. The final prediction is the average of softmax scores of all clips.
We explored using extremely long sequence and its effect on final classification accuracy. 
The classification accuracy saturates at a sequence length of 24. Given the 15fps frame rate in MSRAction3D, a sequence length of 24, i.e. 1.6s, is close to the average length of a complete action. 

\begin{figure}[h]
\small
\centering
\includegraphics[width=\linewidth]{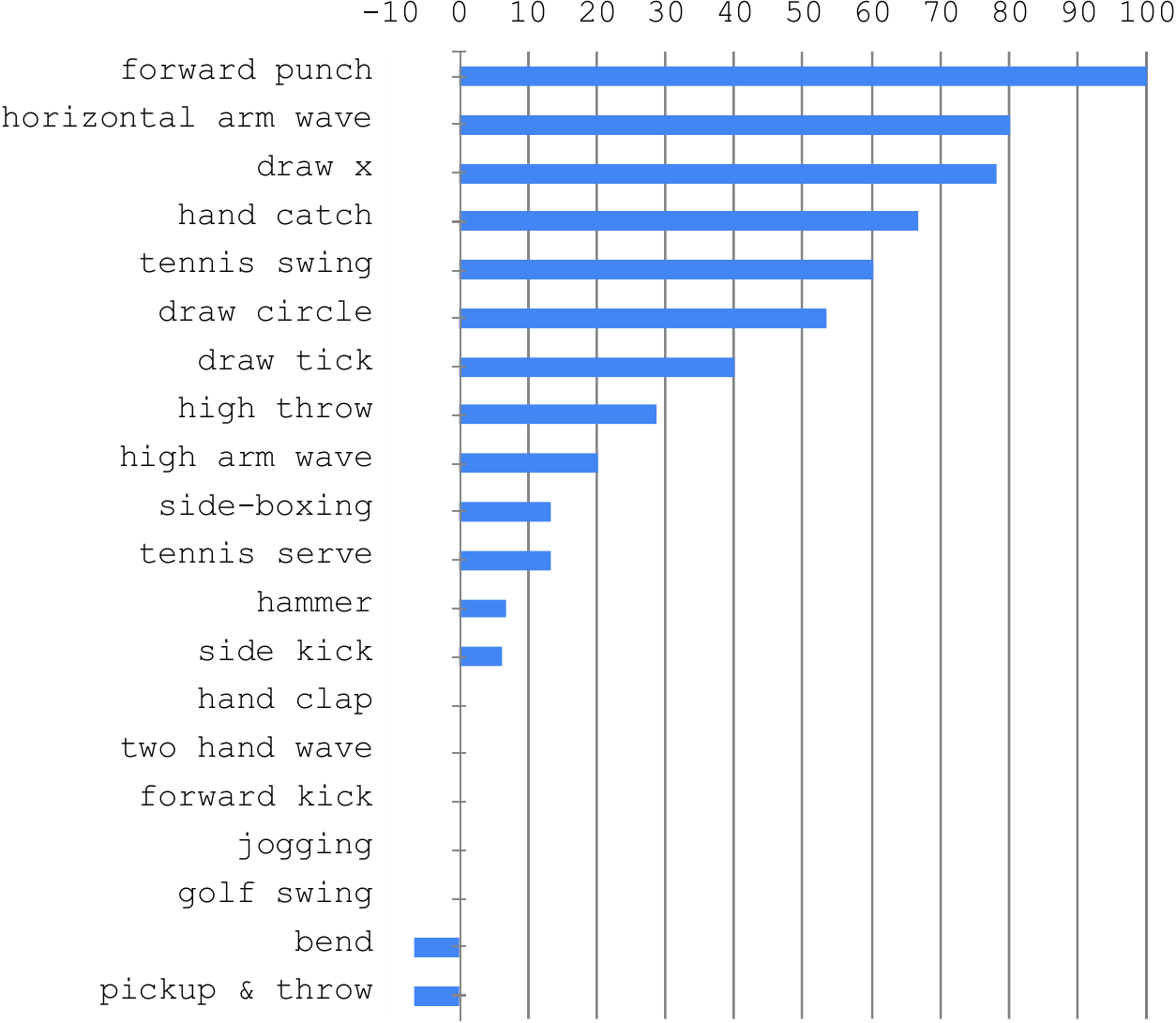}
\caption{ Per-class accuracy gain (\%) of 24 frames MeteorNet-cls  compared to PointNet++. }
\label{fig:msr:acc:gain}
\vspace{-1ex}
\end{figure}

In Figure \ref{fig:msr:acc:gain}, we illustrate the per-class accuracy gain of MeteorNet-cls with 24 frames as input compared to PointNet++ with 1 frame as input. 

We can see that categories that may only be discriminated when observed over time show a significant gain in accuracy when using a sequence of point clouds as input. Categories that can be easily discriminated without temporal information show little or a negative gain in accuracy. For example, the categories ``forward punch'', ``horizontal arm wave'' and ``draw x'' show a large improvement in accuracy. These three categories are similar since they all involve stretching arms forward and thus requires temporal information to be correctly classified. Categories such as “pick up \& throw” or “golf waving” have a very discriminative posture even in single frames and therefore show only the slightest or negative accuracy gain.

The results support our intuition that the Meteor module effectively captures dynamic content of point cloud sequences.

\begin{figure*}
    \centering
    \begin{subfigure}[b]{\linewidth}
        \hspace{6ex}
        \includegraphics[height=0.12\linewidth]{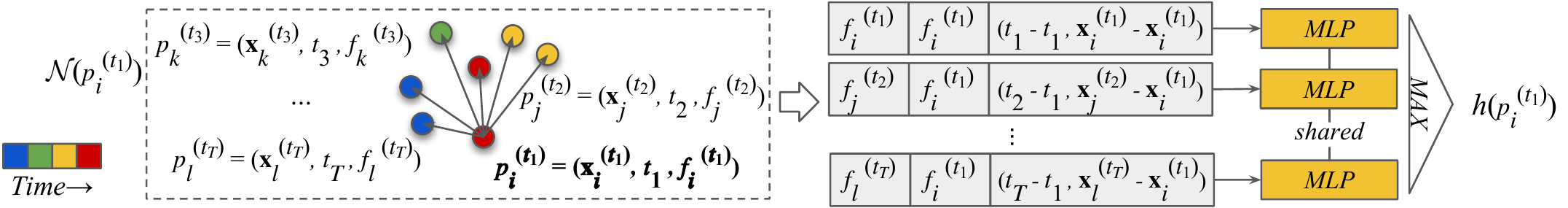}
        \caption{}
        \label{fig:meteor:rel}
    \end{subfigure}
    \begin{subfigure}[b]{\linewidth}
        \hspace{6ex}
        \includegraphics[height=0.12\linewidth]{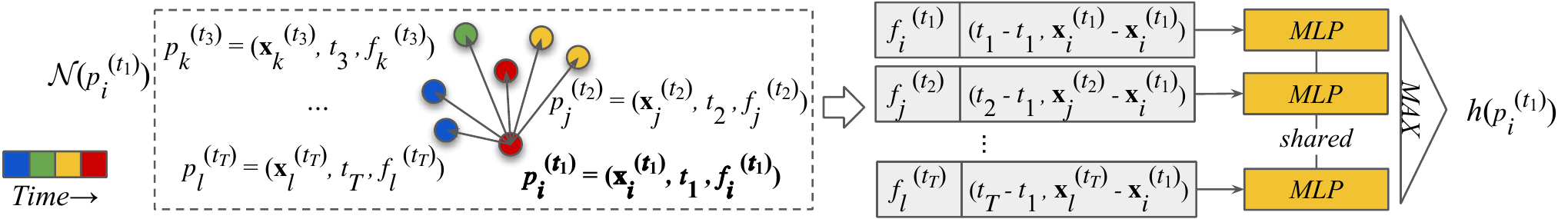}
        \caption{}
        \label{fig:meteor:ind}
    \end{subfigure}
    \caption{ The architecture of (a) \textbf{Meteor-rel module} and (b) \textbf{Meteor-ind module}. 
The dashed box denotes the neighborhood $\mathcal{N}(p_i^{(t)})$ of $p_i^{(t)}$ (in bold) from which all arrows start. The neighborhood $\mathcal{N}$ can be determined by direct grouping or chained-flow grouping. In the figure, $\mathbf{x}$, $t$ and $f$  denotes the 3D spatial coordinate, time coordinate and feature vector of a point respectively;
``MLP'' denotes $\zeta$ in Equation \eqref{eq:rel} and \eqref{eq:ind}, which is the multi-layer individually and independently perceptron applied.
}
    \label{fig:meteor:modules}
\end{figure*}

\section{Outlier Ratio of Scene flow}
\label{sec:scene:flow:outlier}

The ratio of outliers is an important metric that evaluates the robustness of scene flow estimation. We investigate scene flow outlier ratio on KITTI scene flow dataset \cite{KITTI:Scene:flow}. We set different EPE threshold for determining outliers and list the outlier ratio in Table \ref{tab:kitti:flow:more}.

\begin{table}[t]
\small
\setlength{\tabcolsep}{1pt}
\centering
\begin{tabular}{c|c|c|c|c|c|c|c|c|c}
\hline
\multirow{2}{*}{Method}   & \multirow{2}{*}{Frames} & \multicolumn{7}{c}{Threshold for  outlier (m)}  \\ \cline{3-10}
& & 0.3 & 0.4 & 0.5 & 0.6 & 0.7 & 0.8 & 0.9 & 1.0 \\ \hline
FlowNet3D \cite{FlowNet3D} & 2 & 6.88 & 4.28 & 2.19 & 1.31 & 1.02 & 0.77 & 0.59 & 0.54 \\ \hline
\multirow{2}{*}{\begin{tabular}[c]{@{}c@{}}\textbf{MeteorNet-flow} \\ \textbf{(direct)} \end{tabular}}  & 3 & 8.43 & 3.71 & 1.87 & 1.12 & \textbf{0.82} & \textbf{0.65} & 0.55 & \textbf{0.42} \\ 
& 4 & 6.67 & \textbf{3.32} & \textbf{1.44} & \textbf{1.09} & 0.84 & \textbf{0.65} & \textbf{0.53} & \textbf{0.42} \\ \hline
\multirow{2}{*}{\begin{tabular}[c]{@{}c@{}}\textbf{MeteorNet-flow} \\ \textbf{(chain)} \end{tabular}}  & 3 & \textbf{6.35} & 3.72 & 2.39 & 1.49 & 1.16 & 0.92 & 0.74 & 0.62 \\ 
& 4 & 7.88 & 3.50 & 1.95 & 1.27 & 0.85 & 0.72 & 0.63 & 0.58 \\ \hline
\end{tabular}
\caption{Scene flow EPE outlier ratio (\%) given different threshold values. 
}
\label{tab:kitti:flow:more}
\end{table}

As we can see, with more frames as input, MeteorNet-flow can reduce outlier ratio over FlowNet3D.
Besides, MeteorNet-flow using chained-flow grouping with 3 frames as input has the best outlier ratio for a small threshold. However, when the threshold gets larger , MeteorNet-flow using direct grouping is advantageous.

\section{Architecture Details}
\label{sec:arch:detail}

In this section, we provide details on the architectures used in the main paper. We used the same notation as the main paper and assume the input point cloud sequence is $( \{p_i^{(1)}\}, \ldots,  \{p_i^{(T)}\})  \in \mathcal{X}_1 \times \mathcal{X}_2 \times \ldots \times \mathcal{X}_T$ and the local spatiotemporal neighborhood of $p_i^{(t)}$ is $\mathcal{N}(p_i^{(t)})$. 

\subsection{Meteor Module Architecture}
For every point $p_i^{(t)}$ in the point cloud sequence $\{p_i^{(t)}\}$, Meteor module calculates its updated feature vector 
$h(p_i^{(t)})$. In Section 3.2 of the main paper,  we presented two instantiation of $h$.

The first instantiation is for applications where point correspondence is important, such as scene flow. For each ($p_j^{(t')}, p_i^{(t)})$ pair,  we pass the feature vectors of two points and their 4D position difference into to an MLP with shared weights $\zeta$, followed by an element-wise max pooling
\begin{equation} \label{eq:rel}
h(p_i^{(t)}) = \underset{p_j^{(t')}\in \mathcal{N}(p_i^{(t)})}{MAX} \{ \zeta(f_j^{(t')}, f_i^{(t)}, \mathbf{x}_j^{(t')} - \mathbf{x}_i^{(t)}, t'-t) \}
\end{equation}
This instantiation is able to learn the relation between two frames of point clouds. We name the resulting Meteor module \emph{Meteor-rel}. The architecture of Meteor-rel is illustrated in Figure \ref{fig:meteor:modules}(\subref{fig:meteor:rel}).

The second instantiation is for applications where point correspondence is not important, such as semantic segmentation. We pass the feature vector of $p_j^{(t')}$ and 4D position difference between  $p_j^{(t')}$ and  $p_i^{(t)}$ to $\zeta$ followed by an element-wise max pooling 
\begin{equation} \label{eq:ind}
h(p_i^{(t)}) = \underset{p_j^{(t')}\in \mathcal{N}(p_i^{(t)})}{MAX} \{ \zeta(f_j^{(t')}, \mathbf{x}_j^{(t')} - \mathbf{x}_i^{(t)}, t'-t) \}
\end{equation}
We name the resulting Meteor module \emph{Meteor-ind}. Its architecture is illustrated in Figure \ref{fig:meteor:modules}(\subref{fig:meteor:ind}).

Similar to pooling in CNN, the output of both Meteor-ind and Meteor-rel modules can be downsampled by farthest-point-sampling. 

\subsection{MeteorNet-cls Architecture}
\begin{figure*}[t]
\small
\includegraphics[width=\linewidth]{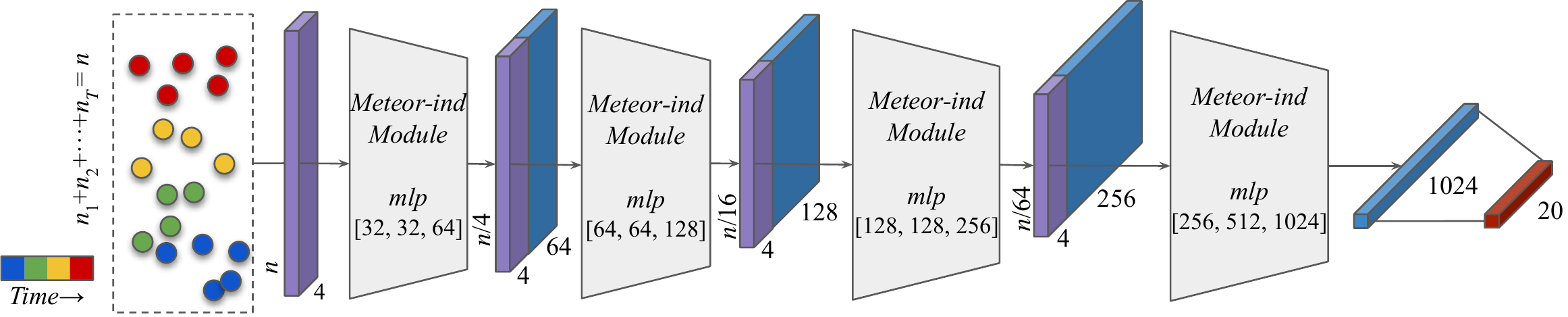}
\caption{ The architecture of \textbf{MeteorNet-cls}. 
}
\label{fig:meteor:cls}
\end{figure*}

MeteorNet-cls $\mathcal{C}$ takes a point cloud sequence  $\{p_i^{(t)}\}$ as input and produces a classification score $c$ for the whole sequence
$$
c = \mathcal{C} (\{p_i^{(1)}\}, \{p_i^{(2)}\}, \ldots, \{p_i^{(T)}\})
$$
MeteorNet-cls consists of four \textbf{Meteor-ind} modules and used \textbf{Early fusion} where the points from different frames are mixed at the first layer. The final Meteor-ind module will max-pool the point cloud to be only one point. The final fully-connected (FC) layer is 20 dimensional which corresponds to the number of classes in the MSRAction3D dataset. The final FC layer is deployed with a dropout layer with dropout rate of 0.5 for regularization. The architecture of MeteorNet-cls is illustrated in Figure \ref{fig:meteor:cls}.

\subsection{MeteorNet-seg Architecture}

\begin{figure*}[t]
\small
\includegraphics[width=\linewidth]{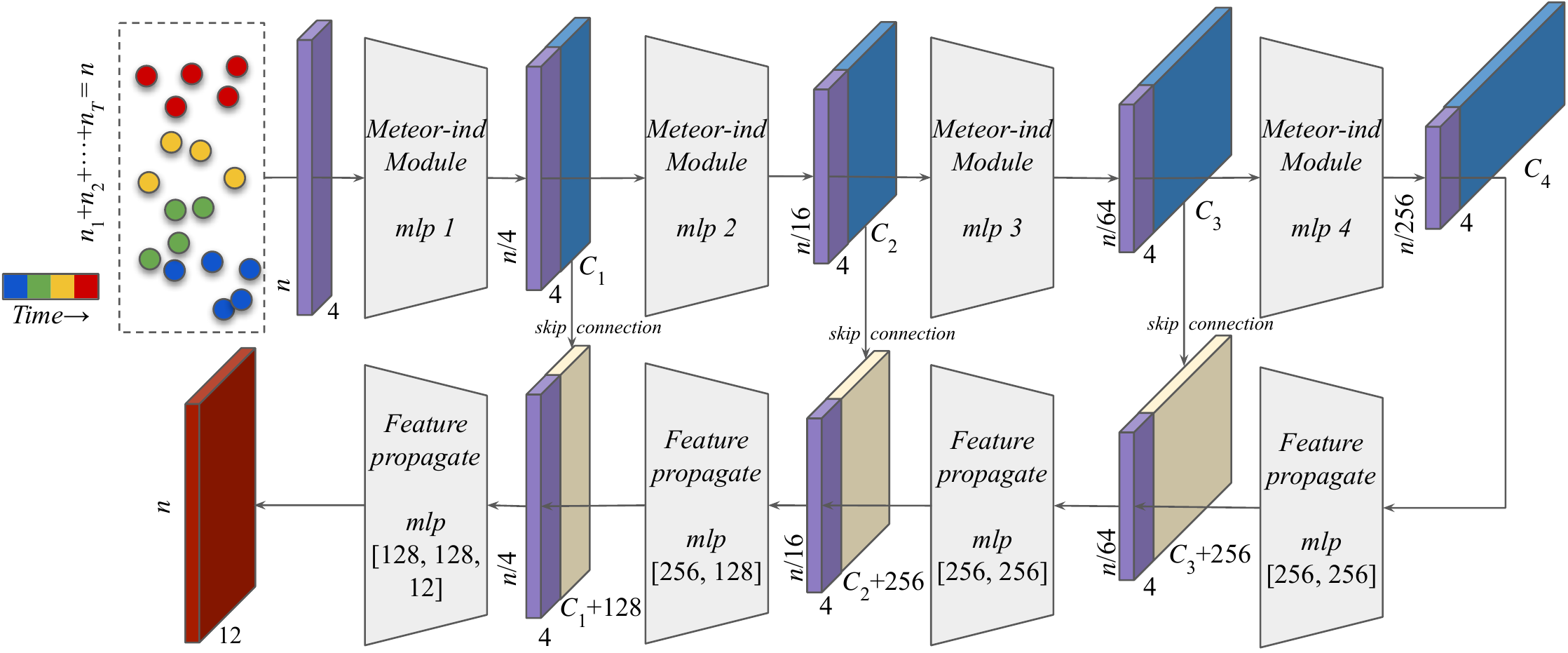}
\caption{ The architecture of \textbf{MeteorNet-seg}. The widths of ``mlp \{1,2,3,4\}'' for different configurations are listed in Table \ref{tab:meteornet:seg:specs}.
}
\label{fig:meteor:seg}
\end{figure*}

MeteorNet-seg $\mathcal{S}$ takes a point cloud sequence  $\{p_i^{(t)}\}$ as input and produces a classification score $c_i^{(t)}$ for every point in the sequence
$$
(\{c_i^{(1)}\}, \ldots, \{c_i^{(T)}\}) = \mathcal{S} (\{p_i^{(1)}\}, \{p_i^{(2)}\}, \ldots, \{p_i^{(T)}\})
$$
MeteorNet-seg consists of four \textbf{Meteor-ind} modules and used \textbf{Early fusion} where the points from different frames are mixed at the first layer.  The point cloud will first be downsampled and then upsampled to the original point cloud through feature propagation layers \cite{PointNet++}. We added skip connections so that local features at early stages of the network can be used in the feature propagation. The output has 12 channels, same number as the number of classes in the Synthia dataset. The final FC layer is deployed with a dropout layer with dropout rate of 0.5 for regularization.
The architecture of MeteorNet-seg is illustrated in Figure \ref{fig:meteor:seg}.

\begin{table}[t]
\small
\setlength{\tabcolsep}{3.5pt}
\centering
\begin{tabular}{l|c|c|c}
\hline
& MeteorNet-seg-$s$ & MeteorNet-seg-$m$ & MeteorNet-seg-$l$ \\ \hline
mlp 1 & [32,32,64] & [32,32,128] & [32,64,128] \\
mlp 2 & [64,64,128] & [64,64,256] & [64,128,256] \\
mlp 3 & [128,128,256] & [128,128,512] & [128,256,512] \\
mlp 4 & [256,256,512] & [256,256,1024] & [256,512,1024] \\
\hline
\end{tabular}
\caption{Architecture configuration for different versions of MeteorNet-seg. ``mlp \{1,2,3,4\}'' corresponds to MLPs of Meteor modules in Figure \ref{fig:meteor:seg}.
}
\label{tab:meteornet:seg:specs}
\end{table}

An ablation study in Section 5.2 of the main paper explored several architecture choices. We listed the architecture configurations in Table \ref{tab:meteornet:seg:specs}. Compared to MeteorNet-seg-$s$, MeteorNet-seg-$m$ has a larger bottleneck dimension at each max pooling layer. Compared to MeteorNet-seg-$m$, MeteorNet-seg-$l$ has the same max pooling dimensions but larger dimensions in non-bottleneck layers.

\subsection{MeteorNet-flow Architecture}

MeteorNet-flow $\mathcal{F}$ takes a point cloud sequence  $\{p_i^{(t)}\}$ as input and estimates a flow vector $\mathfrak{f}_i^{(T)}$ for every point in frame $T$
$$
\{\mathfrak{f}_i^{(T)}\} = \mathcal{F} (\{p_i^{(1)}\}, \{p_i^{(2)}\}, \ldots, \{p_i^{(T)}\})
$$
MeteorNet-flow used \textbf{Late fusion}.
It first employs per-frame set abstraction layers \cite{PointNet++} to downsample the point clouds and learn local features for each frame individually. Then, one \textbf{Meteor-rel} module is used to aggregate information from all frames. Only the points in frame $T$ are selected for subsequent part of the network.
After further processing with feature propagation, MeteorNet-flow obtains the per-point flow vector for every point in frame $T$. We added skip connections so that local features at early stages of the network can be used in feature propagation. The architecture of MeteorNet-flow is illustrated in Figure \ref{fig:meteor:flow}.

\begin{figure}[h]
\small
\includegraphics[width=\linewidth]{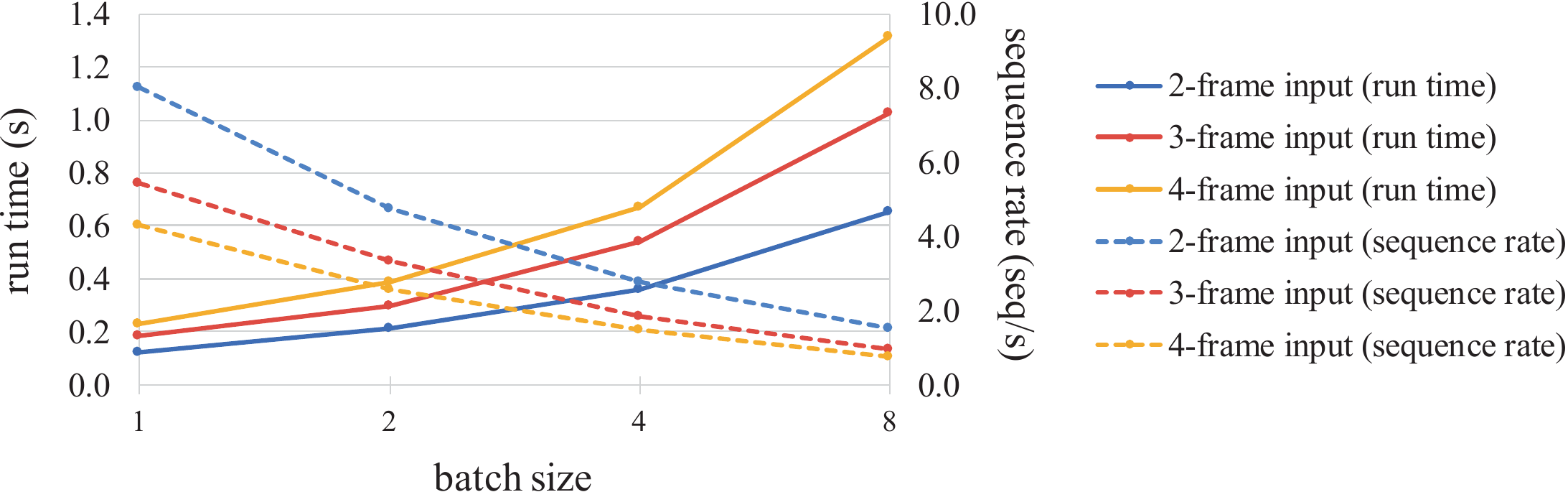}
\caption{ Run time and frame rate of Meteor-seg with direct grouping on Synthia test set. 
}
\label{fig:runtime:direct}
\end{figure}

\begin{figure}[h]
\small
\includegraphics[width=\linewidth]{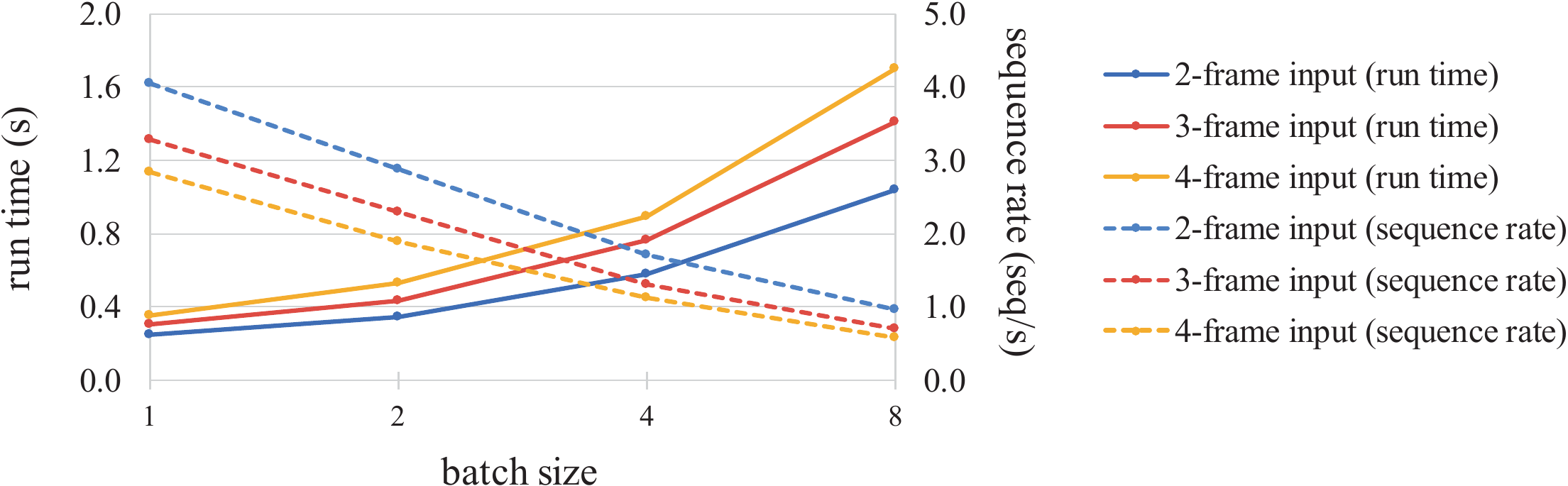}
\caption{ Run time and frame rate of Meteor-seg with chained-flow grouping on Synthia test set. 
}
\label{fig:runtime:chain}
\end{figure}

\section{Model Run Time Analysis}
\label{sec:model:runtime}

We use MeteorNet-seg on the Synthia semantic segmentation test set for runtime analysis. We tested with  8,192 points for the whole scene in each frame. We used a single GTX 1080 Ti GPU and Intel Core i7 CPU. The deep learning framework is Tensorflow 1.9.0. We performed a grid search over batch size and number of frames. The results for direct grouping and chained-flow grouping are illustrated in Figure \ref{fig:runtime:direct} and Figure \ref{fig:runtime:chain} respectively.

Interpolating and chaining flow introduces an additional computational overhead for chained-flow grouping. For 2 frames and a batch size of 1, MeteorNet-seg with direct grouping runs at 8.0 sequences per second (seq/s); MeteorNet-seg with chained-flow grouping runs at 4.1 seq/s.
 For 4 frames and batch size of 1, MeteorNet-seg with direct grouping runs at 4.3 seq/s; MeteorNet-seg with chained-flow grouping runs at 2.8 seq/s.

\section{Proof of Theorem}
\label{sec:theorem}

Suppose $\forall t, \mathcal{X}_t = \{S_t \mid S_t\subseteq [0, 1]^m, |S_t| = n, n\in\mathbb{Z}^+ \}$ is the set of $m$-dimensional point clouds inside an $m$-dimensional unit cube at time $t \in \mathbb{Z}$. We define single-frame Hausdorff distance $d_H(S_i,S_j)$ for $S_i\in\mathcal{X}_i$ and $S_j\in\mathcal{X}_j$. $\mathcal{X}=\mathcal{X}_1 \times \mathcal{X}_2 \times \ldots \times \mathcal{X}_T$ is the set of point cloud sequences of length $T$. Suppose $f: \mathcal{X} \rightarrow \mathbb{R}$ is a continuous function on $\mathcal{X}$ w.r.t $d_{seq}(\cdot,\cdot)$, i.e. $\forall \epsilon > 0$, $\exists\delta>0$, for any $S, S^\prime \in \mathcal{X}$, if $d_{seq}(S, S^\prime) < \delta$, $|f(S)-f(S^\prime)| < \epsilon$. Here, we define the distance of point cloud sequences $d_{seq}(\cdot, \cdot)$ as the maximum per-frame Hausdorff distance among all respective frame pairs, i.e. $d_{seq}(S,S^\prime)=\max_t\{d_H(S_t, S_t^\prime)\}$. 
Our theorem says that $f$ can be approximated arbitrarily closely by a large-enough neural network and a max pooling layer with enough neurons.

We first have the following lemma from the supplementary material of \cite{PointNet}, which ensures the universal approximation potential of PointNet.

\begin{lemma} \label{lemma:pointnet}
Suppose  $f: \mathcal{X} \rightarrow \mathbb{R}$ is a continuous set function w.r.t Hausdorff distance $d_H(\cdot, \cdot)$. $\forall \epsilon > 0$, $\exists$ continuous function $\eta$ and $\gamma$
such that for any $S \in \mathcal{X}$, 
$$
\left|f(S) - \gamma \circ \Big(\underset{x\in S}{MAX}\{\eta(x)\} \Big) \right| < \epsilon
$$
where $MAX$ is a vector max operator that takes a set of vectors as input and returns a new vector of the element-wise maximum.
\end{lemma}

Our theorem is proved based on Lemma \ref{lemma:pointnet}. The core idea is that we can map the point cloud sequence indexed by $t$ into the single point cloud space.

\begin{theorem}
Suppose  $f: \mathcal{X}_1 \times \mathcal{X}_2 \times \ldots \times \mathcal{X}_T \rightarrow \mathbb{R}$ is a continuous function w.r.t  $d_{seq}(\cdot, \cdot)$. $\forall \epsilon > 0$, $\exists$ a continuous function $\zeta(\cdot, \cdot)$ and a continuous function $\gamma$,
such that for any $S=(S_1, S_2, \ldots, S_T) \in \mathcal{X}_1 \times \mathcal{X}_2 \times \ldots \times \mathcal{X}_T$, 
$$
\left|f(S) - \gamma \circ \Big(\underset{\mathbf{x}_i^{(t)}\in S_t, t\in \{1,2,\ldots,T\}}{MAX}\{\zeta(\mathbf{x}_i^{(t)}, t)\} \Big) \right| < \epsilon
$$
where $MAX$ is a vector max operator that takes a set of vectors as input and returns a new vector of the element-wise maximum.
\end{theorem}

\begin{proof}
It suffices to prove for $m=1$. 

In the following proof, we use plain $x_i$ instead of bold $\mathbf{x}_i$ to represent scalar value instead of a 3-D vector.

Let $\mathcal{T} = \{S \mid S\subseteq [0, 1], |S| = n\}$. Define function $\psi: \mathcal{X}\rightarrow \mathcal{T}$ as 
$$\psi (S_1, \ldots, S_T)=\{ p_T(x_{i_t},t) \mid x_{i_t}\in S_t, t\in \{1,\ldots,T\} \}$$
where $p_T(x,t)=\frac{x+t-1}{T}$ is a function that maps each of the $T$ $[0,1]$ intervals into a unique place inside $[0,1]$ interval. Notice that $\mathcal{T}=\mathcal{X}_t$, so $d_H$ can also be defined on $\mathcal{T}$.

For any $S \in \mathcal{X}$, $\forall \epsilon' > 0$,  $\exists \delta = \epsilon' T$, such that $\forall S', d_{seq}(S, S') < \delta$, we have
\begin{equation*}
\begin{split}
& d_H(\psi(S), \psi(S')) \\
&=d_H(\psi(S_1,\ldots,S_T), \psi(S'_1,\ldots,S'_T)) \\
&= \max_{t} \{\sup_{x\in S_t}\inf_{y\in S'_t} d(p_T(x,t), p_T(y,t)), \\
& \sup_{y\in S'_t}\inf_{x\in S_t} d(p_T(y,t), p_T(x,t)) \} \\
& = \max_{t} \{\sup_{x\in S_t}\inf_{y\in S'_t} \frac{1}{T} d(x, y), \sup_{y\in S'_t}\inf_{x\in S_t} \frac{1}{T} d(y, x) \} \\
& = \frac{1}{T} \max_{t} \{\sup_{x\in S_t}\inf_{y\in S'_t} d(x, y), \sup_{y\in S'_t}\inf_{x\in S_t} d(y, x) \} \\
& = \frac{1}{T} \max_t d_H(S_t, S'_t) = \frac{1}{T} d_{seq}(S, S') < \frac{1}{T} \delta = \epsilon'
\end{split}
\end{equation*}
So $\psi$ is a continuous function w.r.t. $d_H: \mathcal{X}\rightarrow \mathbb{R}$ and $d_{seq}: \mathcal{X}'\rightarrow \mathbb{R}$. It's easy to show that the inverse of $\psi$ is also a continuous function.

\begin{figure*}[t]
\small
\centering
\includegraphics[width=\linewidth]{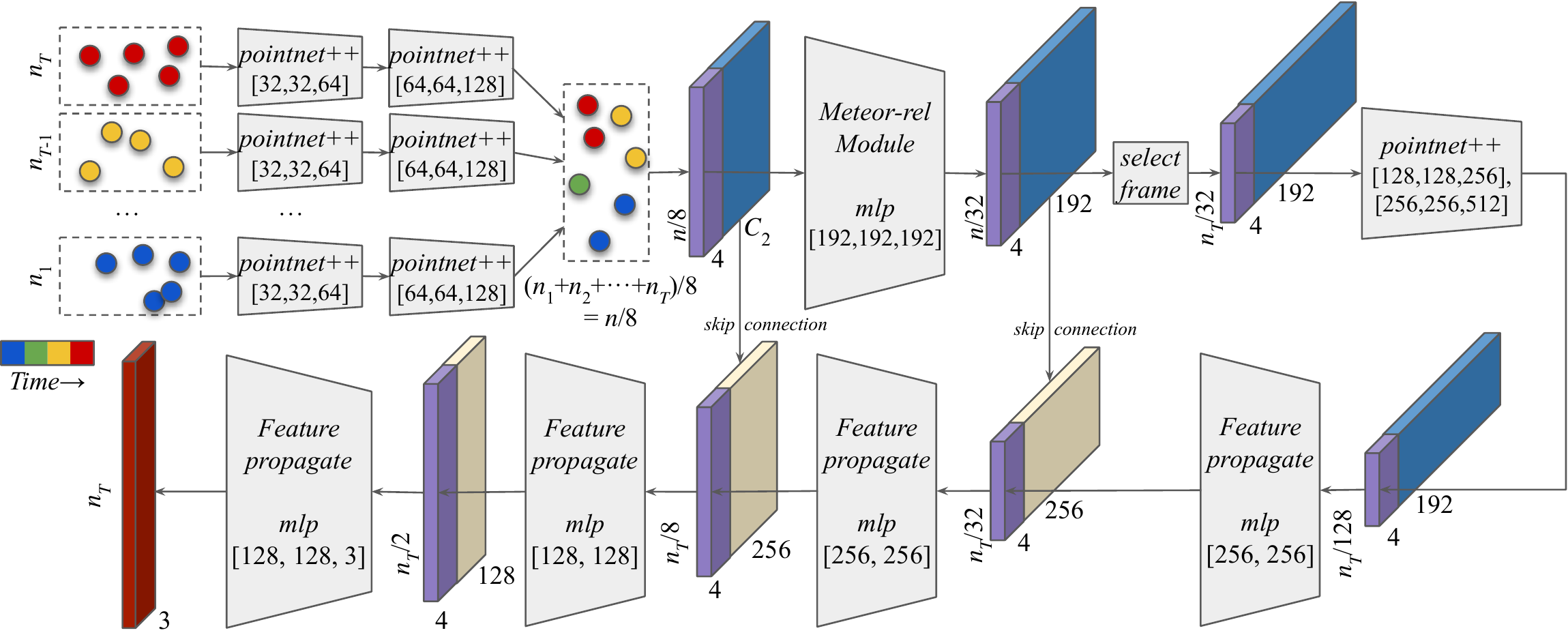}
\caption{ The architecture of \textbf{MeteorNet-flow}. 
}
\label{fig:meteor:flow}
\end{figure*}

According to Lemma \ref{lemma:pointnet}, $\forall\epsilon > 0$, $\exists$ continuous function $\eta$ and $\gamma$
such that for any $\psi(S) \in \mathcal{T}$, 
\begin{equation*}
\begin{split}
& \left|f(S) - \gamma \circ \Big(\underset{x\in \psi(S)}{MAX}\{\eta(x)\} \Big) \right| \\
& = \left|f(S) - \gamma \circ \Big(\underset{x_{i_t}\in S_t, t\in \{1,\ldots,T\}}{MAX}\{\eta(p_T(x_{i_t},t))\} \Big) \right| \\
&= \left|f(S) - \gamma \circ \Big(\underset{x_{i_t}\in S_t, t\in \{1,\ldots,T\}}{MAX}\{\zeta(x_{i_t},t)\} \Big) \right| < \epsilon
\end{split}
\end{equation*}
where $\zeta$ is defined as $\zeta(\cdot, t) = \eta(p_T(\cdot, t))$.

This concludes the proof.
\end{proof}

\section{More Visualization}
\label{sec:viz}


\subsection{Synthia}

\begin{figure}[h]
\small
\includegraphics[width=\linewidth]{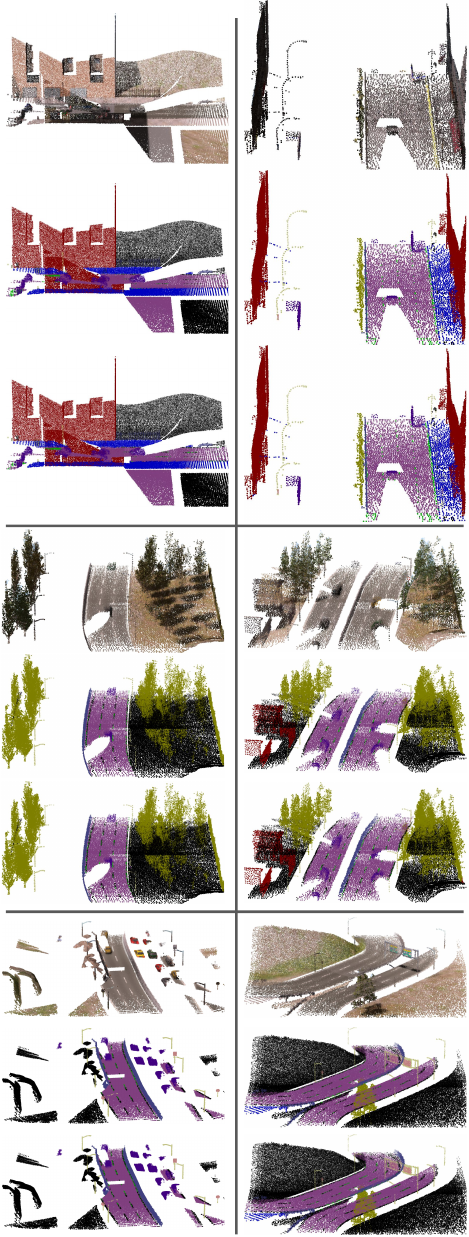}
\caption{  Visualization of two example results from the Synthia dataset. Each cell from the top: RGB input, ground truth, predictions.
}
\label{fig:synthia}
\end{figure}

We provide additional qualitative results for segmentation results on the Synthia test set in Figure \ref{fig:synthia}. Again, MeteorNet-seg can accurately segment most objects.

\begin{figure}[h]
\small
\includegraphics[width=\linewidth]{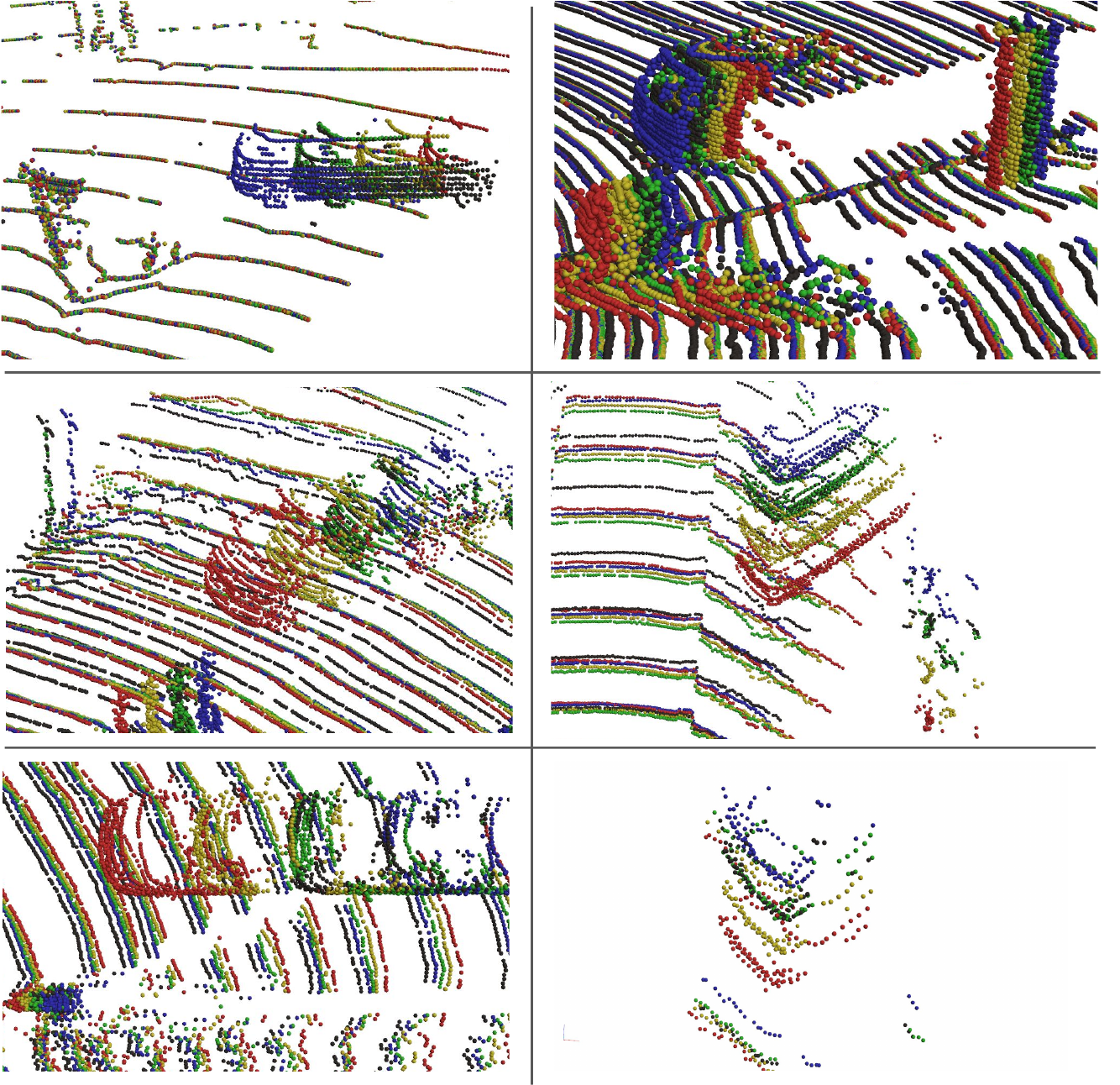}
\caption{ Additional visualization of MeteorNet example results on the KITTI scene flow dataset. Point are colored to indicate which frames they belong to: \textcolor{blue}{frames $t-3$}, \textcolor{viz_green}{frame $t-2$}, \textcolor{viz_yellow}{frame $t-1$}, \textcolor{viz_red}{frame $t$}. 
\textbf{Translated points} (frame $t-3$ + estimated scene flow) is in black. Green and black shapes are supposed to overlap for perfect estimation.
}
\label{fig:kitti:flow}
\end{figure}

\subsection{KITTI scene flow}

We provide additional qualitative results for scene flow estimation results on KITTI scene flow dataset in Figure \ref{fig:kitti:flow}. Again, MeteorNet-flow can accurately estimate flow for moving objects.

\section{Name Metaphor}
\label{sec:metaphor}

The universe is all of space and time. When we look deep into the universe, stars are the visible points in the sky. Meteor shower is a group of stars that move together as a ``dynamic point cloud sequence''. It brings fortune and good luck to anyone who sees it. 

We hope our MeteorNet can also bring fortune and good luck to our readers and benefit related research domains.

\section{Changelog}

Feb 10th: Fixed the last column of Table \ref{tab:synthia:seg} with mean accuracy results.

\end{document}